\documentclass[10pt,twocolumn,letterpaper]{article}

\usepackage[pagenumbers]{cvpr} 
\usepackage{multirow}
\usepackage{algorithm}
\usepackage{algorithmic}
\usepackage{makecell}
\usepackage{bbding}
\usepackage[T1]{fontenc}

\definecolor{cvprblue}{rgb}{0.21,0.49,0.74}
\usepackage[pagebackref,breaklinks,colorlinks,allcolors=cvprblue]{hyperref}

\def\paperID{***} 
\def\confName{***}
\def\confYear{***}

\title{HIMOSA: Efficient Remote Sensing Image Super-Resolution with \underline{Hi}erarchical \underline{M}ixture \underline{o}f \underline{S}parse \underline{A}ttention}

\author{
Yi Liu \quad
Yi Wan$^{\dagger}$ \quad
Xinyi Liu \quad
Qiong Wu \quad
Panwang Xia \quad
Xuejun Huang \quad
Yongjun Zhang$^{\dagger}$
\\
Wuhan University \quad 
{$^{\dagger}$ Corresponding author.} \\
{\tt\small \{liuyiwhu28, yi.wan, zhangyj\}@whu.edu.cn}
}

\begin{document}
\maketitle
\begin{abstract}
In remote sensing applications, such as disaster detection and response, real-time efficiency and model lightweighting are of critical importance. Consequently, existing remote sensing image super-resolution methods often face a trade-off between model performance and computational efficiency. In this paper, we propose a lightweight super-resolution framework for remote sensing imagery, named HIMOSA. Specifically, HIMOSA leverages the inherent redundancy in remote sensing imagery and introduces a content-aware sparse attention mechanism, enabling the model to achieve fast inference while maintaining strong reconstruction performance. Furthermore, to effectively leverage the multi-scale repetitive patterns found in remote sensing imagery, we introduce a hierarchical window expansion and reduce the computational complexity by adjusting the sparsity of the attention. Extensive experiments on multiple remote sensing datasets demonstrate that our method achieves state-of-the-art performance while maintaining computational efficiency.
\end{abstract}    
\section{Introduction}
\label{sec:intro}

In recent years, with the rapid advancement of remote sensing technologies and their expanding range of applications, high-resolution remote sensing imagery (RSI) has played an increasingly critical role in fields such as urban planning, agricultural monitoring, change detection, and disaster response. In particular, during natural disasters such as earthquakes, floods, and wildfires, timely access to high-quality imagery is essential for accurate disaster assessment and rapid response, placing strict requirements on data acquisition and processing efficiency. However, due to limitations in onboard sensor capabilities, transmission bandwidth, and complex imaging conditions, directly capturing native high-resolution images often implies high costs and significant delays, which limit their effectiveness in time-sensitive scenarios. 

\begin{figure}[t]
\centering
\includegraphics[width=0.96  \linewidth]{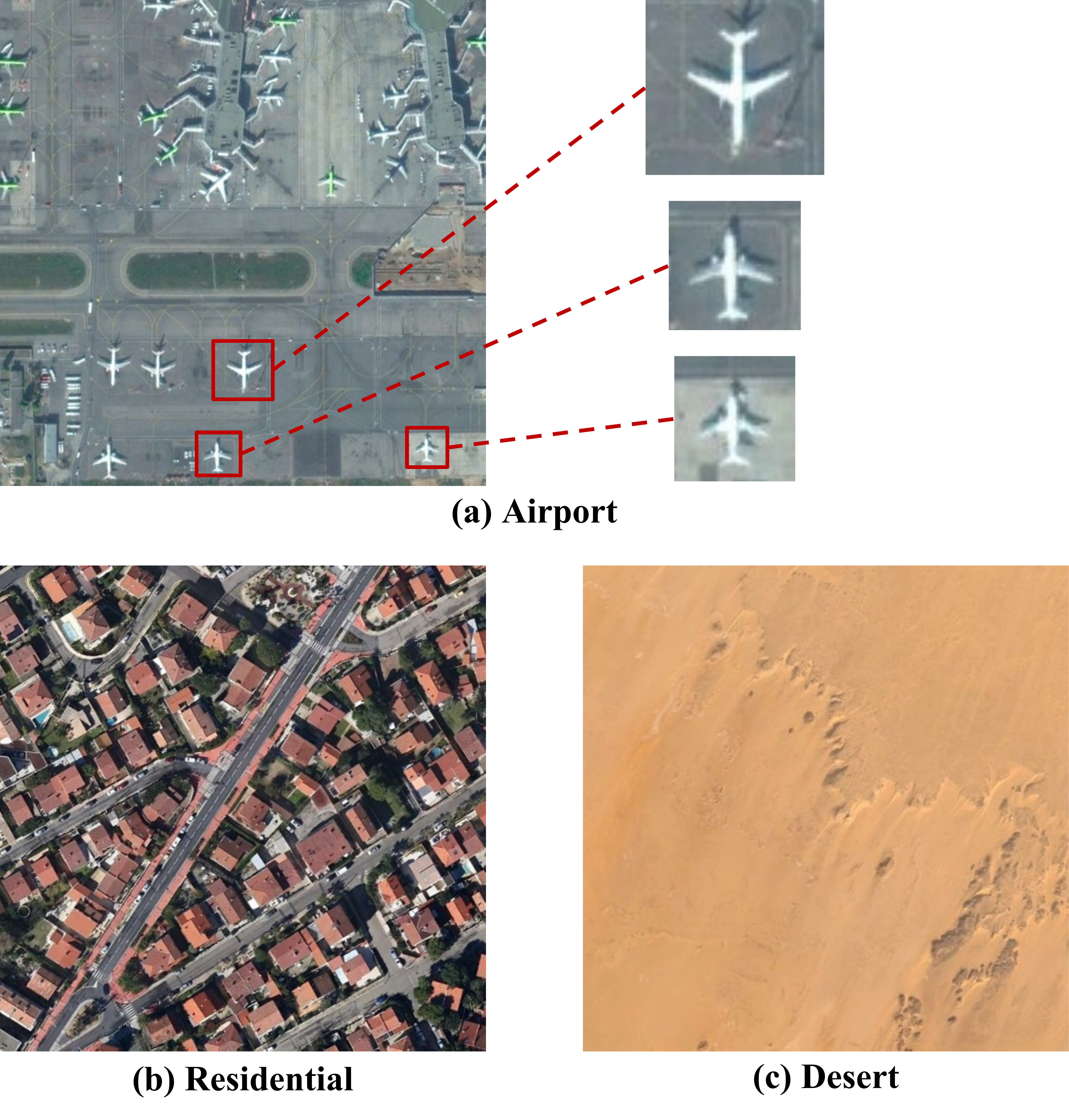}
\vspace{-2mm}
\caption{Different visual patterns in remote sensing imagery. (a) multi-scale repetitive patterns; (b) repetitive patterns; (c) weak texture.}
\vspace{-4mm}
\label{fig.Intro}
\end{figure}

To address this challenge, single-image super-resolution (SISR) has been developed to reconstruct high-resolution images from low-resolution observations, offering a more flexible and cost-effective alternative. As a representative task in low-level vision, SISR is inherently ill-posed. Early approaches attempted to address this challenge by employing upsampling techniques or incorporating handcrafted priors to constrain the solution space. With the rise of deep learning, Convolutional Neural Networks (CNN)~\cite{srcnn,edsr} have achieved remarkable success in super-resolution by modeling local features such as edges and color patterns using small convolutional kernels (\eg, $3\times3$). Although CNN-based methods have advantages in inference efficiency, their limited receptive fields and fixed convolutional kernels restrict global context modeling and adaptability, which are crucial for reconstructing high-resolution details.

Benefiting from powerful self-attention~\cite{transformer} mechanisms, Transformer-based models have demonstrated superior performance over CNN-based methods in super-resolution tasks. These models typically employ dense self-attention to aggregate global features by computing pairwise similarities among all image tokens. However, this approach suffers from quadratic computational complexity $o(N^2)$. To balance computational complexity and long-range modeling capability, some methods ~\cite{liang2021swinir, zhangSwinFIRRevisitingSwinIR2023a} introduced window attention mechanisms, which reduce computational complexity by restricting attention to non-overlapping local windows. While this strategy significantly improves efficiency, it inevitably weakens the ability to model long-range dependencies, especially when the window size is small. Recent studies \cite{chen2023activating} further suggest that enlarging the window size can effectively expand the receptive field and enhance super-resolution performance, reinforcing the importance of capturing long-range dependencies in SR tasks. To address this issue, some methods~\cite{PFTNet, TTST} use sparse attention within a large window to reduce computational complexity, while other methods~\cite{ATDNet, CATANet} use token clustering to model long-range dependencies through latent semantic information. Nevertheless, these methods still require all tokens in the window or cluster to participate in similarity computations, which limits their computational efficiency.

Considering the inherent large-scale Earth observation in remote sensing, remote sensing images (RSIs) exhibit several characteristics that are uncommon in natural imagery. As illustrated in~\cref{fig.Intro}, these include, but are not limited to, multi-scale repetitive patterns (\cref{fig.Intro}(a)(b)) and large amounts of redundant information (\cref{fig.Intro}(c)). The redundancy of information makes the sparse Transformer a natural choice for RSI super-resolution. However, existing sparse Transformer-based methods suffer from two major limitations: (1) they rely on manually designed token selection strategies, such as sparse intervals~\cite{art} or top-$k$~\cite{TTST}. The former lacks attention to image content, while the latter requires all tokens to participate in similarity calculations, increasing computational complexity; and (2) the adoption of fixed window sizes limits their ability to capture multi-scale repetitive patterns effectively. Therefore, resolving the above issues is essential for enhancing the performance of RSI super-resolution.

To this end, we propose a novel efficient super-resolution framework to mitigate the aforementioned issues. To reduce the computational complexity, inspired by the Mixture of Experts (MoE)~\cite{moe}, we propose a content-aware routing sparse attention mechanism for efficient remote sensing image super-resolution. Our approach employs Expert-Choice Routing~\cite{zhou2022mixture} to enable dynamic, content-aware, and head-specific token selection. Specifically, we treat each expert as a head in the traditional dense multi-head self-attention and each expert selects $k$ specific tokens from the input. Unlike previous methods that first aggregate all tokens and subsequently enforce sparsity, our approach selects specific tokens first and then performs information aggregation, reducing the computational complexity from $o(N^2)$ to $o(k^2+N)$. To address the prevalent multi-scale patterns in remote sensing imagery, we introduce a hierarchical window-based attention mechanism. Specifically, we replace the fixed-size windows used in previous methods with progressively enlarged windows. This design enables the network to capture informative multi-scale features and gradually expand the receptive field, thereby enhancing multi-scale and global context modeling capabilities.
We highlight our main contributions as follows:
\begin{itemize}
    \item We propose a novel super-resolution framework, named HIMOSA, boosting SR performance by exploiting multi-scale features and long-range dependencies.
    \item We design a content-aware routing sparse attention mechanism to selectively aggregate effective tokens within each window, enabling the efficient utilization of large window sizes.
    \item Extensive experiments on multiple public datasets demonstrate the superior performance of our method in achieving efficient and effective RSI super-resolution and reconstructing high-resolution details.
\end{itemize}
\section{Related Work}
\label{sec:related_work}

In this section, we introduce the related work, including the single-image super-resolution and mixture of experts.
\subsection{Single image super-resolution}
\noindent\textbf{Transformer-based methods.}
The introduction of Vision Transformer (ViT)~\cite{vit} brought Transformer architectures into the vision domain. Owing to the powerful ability of self-attention to capture long-range dependencies, Transformer-based super-resolution methods~\cite{ipt,transenet,restormer,yoo2023enriched, CATANet, ESTNet, stereoinr} have achieved performance comparable to, and in many cases surpassing, CNN-based approaches~\cite{srcnn,edsr, vdsr, esrgan, rcan, rdn, fsrcnn, espcn, srresnet, srdensenet, drcn, drrn, holistic-attenion, nlsa, hsenet}. While the global modeling capability of self-attention is particularly well-suited for remote sensing imagery with large spatial coverage, it also incurs substantial computational complexity. Inspired by Swin-Transformer~\cite{swintransformer}, SwinIR~\cite{liang2021swinir} restricts self-attention computation within shifted local windows, effectively reducing complexity but at the cost of limiting non-local representation due to the small window size. To alleviate this limitation, several variants have been proposed. For example, ATD~\cite{ATDNet} introduces adaptive token dictionaries with category-based attention to better capture global information; CAT~\cite{cat} adopts rectangular window attention combined with axial shifting to enhance cross-window interaction for improved restoration; HAT~\cite{chen2023activating} enlarges the window size and incorporates channel attention to further strengthen inter-window communication, achieving higher-quality reconstruction. Despite these advances, the fixed-size windows employed in these methods inherently lack adaptability to the multi-scale nature of remote sensing imagery, often leading to suboptimal reconstruction quality. To address this issue, HiT-SRF~\cite{zhang2024hit} adopts progressively enlarged windows to capture multi-scale features. However, it does not explicitly handle redundancy within large windows, resulting in limited gains, while the increased computational complexity associated with large-scale windows remains a critical concern.

\noindent\textbf{Sparse Transformer.}
Dense attention often introduces task-irrelevant features, thereby weakening the extraction of useful information. To address this issue, numerous studies have explored sparse attention mechanisms to reduce computational complexity while focusing on key features. NLSA~\cite{nlsa} combines non-local operations with sparse representation, employing dynamic sparse selection to improve efficiency and robustness; ART~\cite{art} integrates dense and sparse attention to achieve a broader receptive field and enhanced feature representation in image restoration. Another line of work~\cite{wang2022kvt, zhao2019explicit} explicitly filters the most relevant positions in the attention map to suppress redundant information. Building on this idea, methods like TTST~\cite{TTST} and DRSformer~\cite{DRSformer} introduce learnable top-$k$ operators to adaptively retain the most relevant attention values, thus improving restoration performance. PFTNet~\cite{PFTNet} leverages progressive feature aggregation (PFA) to gradually refine feature selection and reduce computational complexity; SeemoRe~\cite{zamfir2024seemore} employs an expert mining strategy with mixed low-rank experts to achieve a lightweight design.

\subsection{Mixture of Experts}
Mixture-of-Experts (MoE)~\cite{moe} integrates multiple specialized networks (experts) with a gating mechanism that routes each input to a small subset of experts, producing outputs as a sparsely weighted combination of selected predictions. A major challenge is load balancing: without explicit constraints, routing tends to overuse a few experts while leaving others inactive, thus limiting model capacity~\cite{shazeer2017outrageously}. Traditional approaches mitigate this with auxiliary balancing losses~\cite{fedus2022switch, lepikhin2020gshard}, while expert-choice routing~\cite{zhou2022mixture} ensures balanced utilization by allowing experts to select tokens instead. MoE has been applied to both feed-forward and attention layers, such as SwitchHead~\cite{csordas2024switchhead}, which reduces the number of attention heads by replacing them with MoE, and MoA~\cite{fu2024moa}, which enhances multi-query attention~\cite{ainslie2023gqa}. In addition, MoSA~\cite{pikekos2025mosa} introduces sparsity into the attention mechanism through a mixture-of-experts design, thereby reducing computational complexity. Inspired by these works, we also adopt a mixture-of-experts design to introduce sparsity into the attention computation, thereby improving inference efficiency.
\section{Methodology}
\label{sec:method}

\begin{figure*}[t]
\centering
\includegraphics[width=0.96  \textwidth]{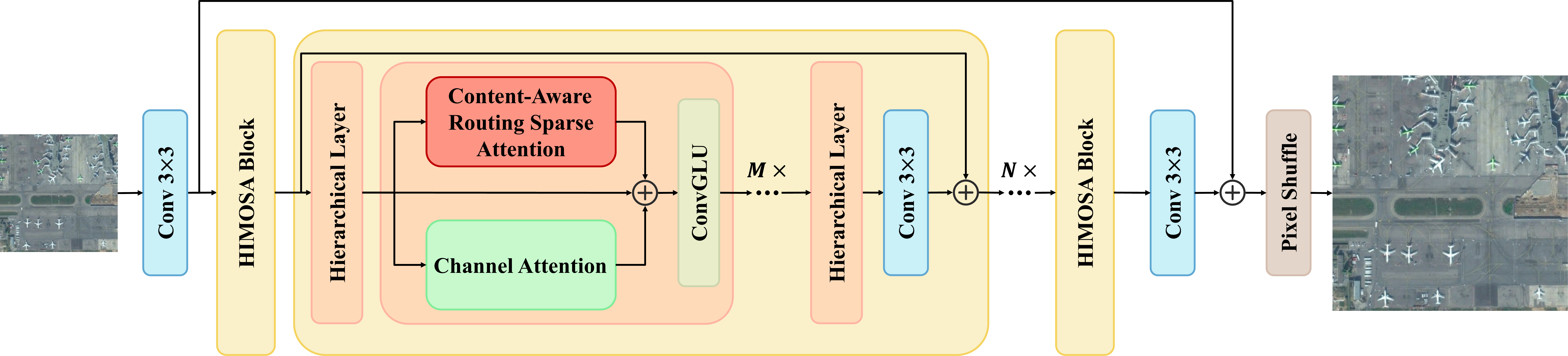}
\vspace{-2mm}
\caption{The overall architecture of HIMOSA. Each HIMOSA block contains $M$ hierarchical layers, each of which includes a content-aware routing sparse attention (CARSA), a channel attention module (CA) and a convolutional gated linear unit (ConvGLU).}
\vspace{-2mm}
\label{fig.Network}
\end{figure*}

\subsection{Overall Architecture}
Given an LR input image $I_{\mathrm{lr}}\in \mathbb{R}^{h\times w\times3}$, we aim to restore the high resolution details and reconstruct the corresponding HR results $I_{\mathrm{sr}}\in \mathbb{R}^{H\times W\times3}$. The overall architecture consists of three main components: shallow feature extraction, deep feature extraction, and image reconstruction, as illustrated in \cref{fig.Network}. 

First, we employ a shallow feature extractor $F$ composed of a $3\times 3$ convolutional layer to obtain low-level features from the input image:
\begin{equation}
    \label{shallow_feature}
    X_{0} = F(I_{\mathrm{lr}}).
\end{equation}

Then, we cascaded $N$ hierarchical mixture of sparse attention (HIMOSA) blocks to progressively extract deep and rich representations from the encoded features. Each HIMOSA block is composed of $M$ hierarchical layers designed to capture both local spatial correlations and long-range contextual dependencies at multiple scales. Within each layer, the content-aware routing sparse attention (CARSA) adaptively selects informative tokens, enabling efficient but expressive token aggregation. The channel attention module (CA) further enhances the global information modeling capability, while the convolutional gated linear unit (ConvGLU) introduces nonlinearity and spatial adaptability to achieve more flexible feature modulation.
Finally, the aggregated hierarchical features are fed into a lightweight reconstruction head that employs PixelShuffle~\cite{pixelshuffle} to upsample the features and generate the final high-resolution output with fine spatial details.

\subsection{Hierarchical Mixture of Sparse Attention}
In this subsection, we introduce the motivation and details of our proposed hierarchical mixture of sparse attention.

In remote sensing imagery, a common characteristic is the presence of multi-scale repetitive patterns. Unlike previous methods that use fixed-scale small windows, we propose progressively expanding the window size, allowing the network to aggregate multi-scale information while increasing its receptive field. However, as the window size increases, the computational complexity grows rapidly. How can we address this challenge? We observe that larger windows tend to contain substantial redundant information, making the attention distribution increasingly sparse. Inspired by recent works such as PFTNet~\cite{PFTNet} and MOSA~\cite{pikekos2025mosa}, we note that not all tokens need to be involved in the similarity computation, especially in a large window, \eg, $64 \times 64$. Therefore, we selectively retain the top-$k$ tokens that contribute most to image reconstruction before performing the similarity computation with quadratic complexity, thereby significantly improving the computational efficiency of our method.

\subsubsection{Hierarchical windows}
Inspired by HiTSR~\cite{zhang2024hit}, we introduce the hierarchical windows to aggregate the multi-scale information. Specifically, given a base window size $\mathit{ws}_{B}$, in each HIMOSA block, we set the window size $\mathit{ws}_{i}$ for the $i$-th hierarchical layer to:
\begin{equation}
    \label{hier-window}
    \mathit{ws}_{i} = \alpha_{i}\mathit{ws}_{B},
\end{equation}
where $\alpha_{i}\in (\alpha_{0},\alpha_{1},..., \alpha_{M})$ denotes the hierarchical ratio for the $i$-th hierarchical layer.

\subsubsection{Content-aware routing sparse attention} In each hierarchical layer, we propose Content-Aware Routing Sparse Attention (CARSA) to efficiently model the relationships among tokens within the hierarchical window. 

Compared with expert-level gating in MOSA~\cite{pikekos2025mosa}, CARSA adopts a content-aware scoring function within each head, ranking tokens by feature similarity and keeping top-$k$ candidates for attention computation. Then, a layer-wise sparsity ratio $\rho_{i}$ controls the number of selected tokens, jointly with hierarchical windows to balance local and global dependencies. This routing scheme introduces deterministic sparsity and better spatial consistency.
Specifically, for each hierarchical layer, the input features are partitioned into non-overlapping windows of size $\mathit{ws}_{i}$.
For notational clarity, we denote the partitioned features of the $i$-th layer as $X_{i} \in \mathbb{R}^{n\times d}$. We introduce a content-aware router to help different experts select tokens. First, we calculate the selection scores $r_{i} \in \mathbb{R}^{n\times m}$ for each token:
\begin{align}
\label{calc_similarity}
    r_{i}=\sigma(X_{i}W^{r})
\end{align}
where $W^{r} \in \mathbb{R}^{d\times m}$ is the linear transform for input tokens, and $\sigma$ is the sigmoid activation function. $m$ denotes the number of experts.

Due to the varying levels of redundancy across different window sizes, processing all tokens in the initial small windows remains computationally affordable. With the expansion of window size, the level of information redundancy within each window correspondingly increases. To this end, we assign different sparsities $\rho_{i}$ to different window sizes, so that the sparsity in the attention calculation gradually increases with the window size. This design allows the model to substantially reduce computational cost while still preserving its ability to capture critical information for high-quality reconstruction. Then we select the top-$k$ indices in the selection score $r_{i}$, which can be formally as:
\begin{align}
    k_{i} & = n/\rho_{i} \\
    r^{topk}_{i}, I_{i} & = \mathrm{TopK}(r_{i}, k_{i})
\end{align}
where $r^{topk}_{i}$ denotes the highest $k$ values of $r_{i}$ and $I_{i}\in\{0,...,n-1\}^{k}$  is the corresponding indices. Then we use $I_{i}$ to select the subset of input tokens for each head:
\begin{align}
    X^{S}_{i} = (X_{I_{1}}, X_{I_{2}},...,X_{I_{k}}) \in \mathbb{R}^{k\times d}
\end{align}

After that, queries, keys, and values are calculated identically to the standard MHA~\cite{transformer}. For each expert, it has its own linear mappings $W^{Q}_{h},W^{K}_{h},W^{V}_{h}\in \mathbb{R}^{d\times d'}$ and the output mapping $W^{O}\in \mathbb{R}^{d'\times d}$, where $h\in[1,m]$:
\begin{align}
    & Q^{S}_{h}=X^{S}_{i}W^{Q}_{h}, K^{S}_{h}=X^{S}_{i}W^{K}_{h}, V^{S}_{h}=X^{S}_{i}W^{V}_{h},\\
    & e_{h} = \mathrm{softmax}(Q^{S}_{h}{K^{S}_{h}}^{T}/\sqrt{d^{'}})V^{S}_{h} \\
    & X_{\mathrm{CARSA}} = \mathrm{Concat}(e_{1},e_{2},...,e_{h})W^{O}
\end{align}

\begin{figure}[t]
\centering
\includegraphics[width=0.96  \linewidth]{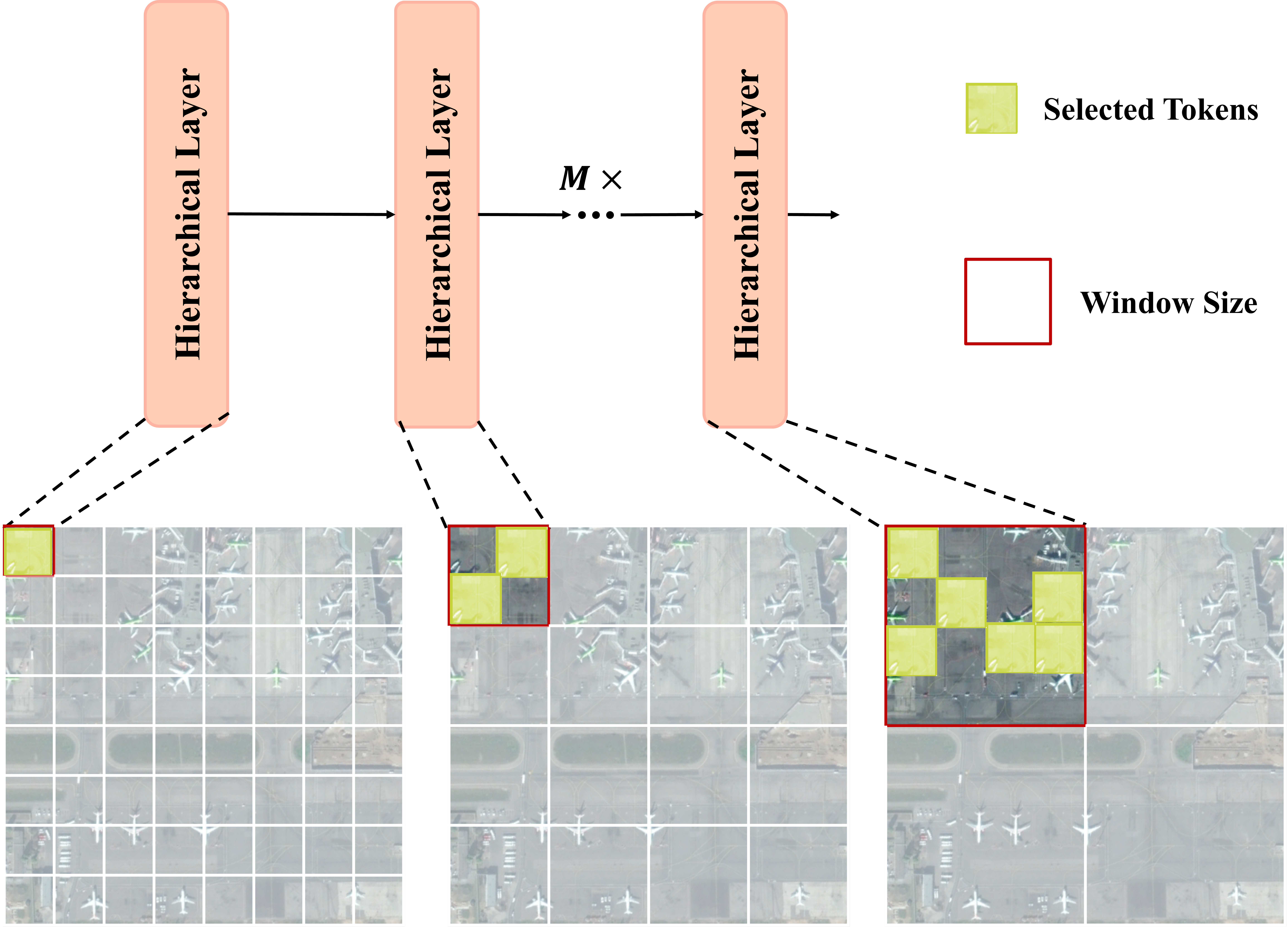}
\vspace{-2mm}
\caption{Hierarchical window for sparse attention. Increasing window sizes are applied to different hierarchical layers to aggregate features with expanding receptive fields.}
\vspace{-2mm}
\label{fig.Hier_windows}
\end{figure}

\begin{figure}[t]
\centering
\includegraphics[width=\linewidth]{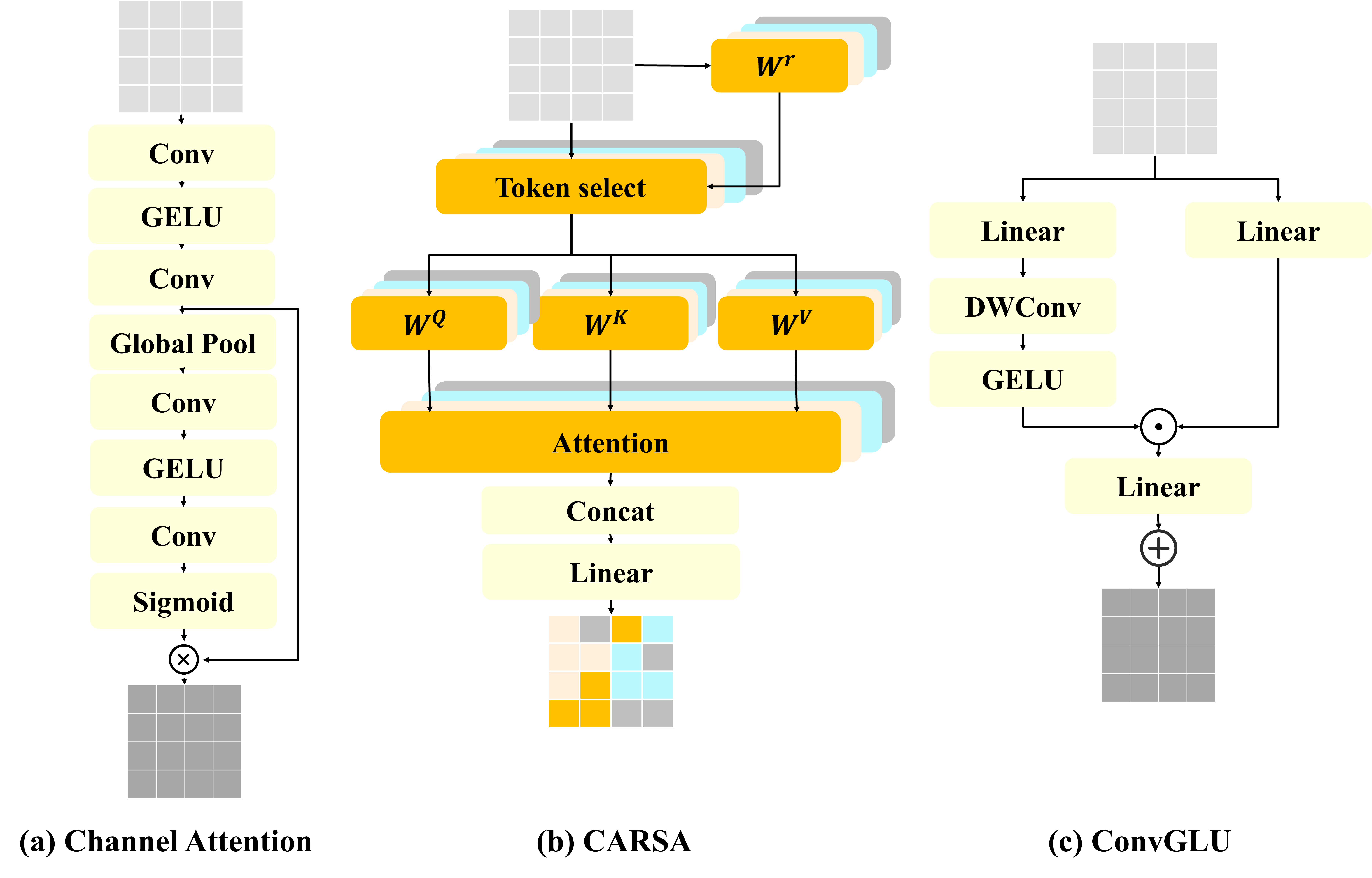}
\vspace{-6mm}
\caption{The structure of: (a) Channel attention; (b) CARSA; (c) ConvGLU.}
\vspace{-4mm}
\label{fig.CARSA}
\end{figure}

\subsubsection{Channel attention}
In conventional expert-choice routing strategies, each expert selects the tokens to process based on routing scores. This “expert-selects-token” mechanism has proven effective in natural language processing (NLP) tasks~\cite{zhou2022mixture}, primarily because semantic information in language data is typically concentrated in a small number of key tokens. 
However, directly applying this strategy to low-level vision tasks such as image super-resolution may lead to performance bottlenecks. Due to the lack of a global coordination mechanism, multiple experts tend to select tokens from semantically salient regions of the image (such as edges or textured regions), resulting in a highly imbalanced token distribution: some tokens are repeatedly processed by multiple experts, while many other tokens (typically located in smooth or low-texture regions) are consistently ignored.
For low-level tasks like super-resolution, this redundant focus on a limited subset of tokens leads to underutilization of image-wide information, thereby reducing reconstruction quality. Therefore, we further incorporate the channel attention block (CAB)~\cite{rcan} to enhance the network’s ability to capture global information. 
 
\subsubsection{ConvGLU} 
Finally, we replaced the standard convolutional feed-forward layer with a convolutional gated linear unit ~\cite{shi2024transnext}. 
As shown in ~\cref{fig.CARSA}, the convolutional GLU consists of two parallel convolutional projections. One branch generates a gating signal through a nonlinear activation function, while the other performs the main feature transformation. The outputs of these two branches are then combined through element-wise multiplication. In this way, the gating branch adaptively controls the contribution of the transformed features, determining which components should be emphasized and which should be suppressed. This adaptive feature selection mechanism allows the model to highlight informative patterns while mitigating irrelevant or redundant signals, significantly improving reconstruction performance without sacrificing computational efficiency.
\section{Experiments and Analysis}
\subsection{Implementation Details}
\noindent\textbf{Datasets.}
In this paper, we employ four remote sensing imagery datasets, including AID~\cite{xia2017aid}, DOTA v2.0~\cite{xia2018dota}, DIOR~\cite{li2020object}, and NWPU-RESISC45~\cite{cheng2017remote}.
For the AID dataset, we randomly select 100 images per scene for training and 30 images per scene for testing, resulting in a total of 3,000 training images and 900 testing images. The NWPU-RESISC45, DIOR, and DOTA v2.0 datasets are used exclusively for testing. 

\noindent\textbf{Training Details.}
There are 4 HIMOSA blocks in total, each comprising 6 hierarchical layers with a channel number of 60. In each HIMOSA block, the base window size $\mathit{ws}_{B}$ is set to 8 and the hierarchical ratio is set to $(0.5, 1, 2, 4, 6, 8)$. For CARSA, the number of experts is set to 8, and the sparsity is set to $(1,1,2,4,8,12)$ as the window expands. During training, we adopt the Muon~\cite{muon} optimizer with an initial learning rate of $5 \times 10^{-4}$. The total number of training iterations is set to 250k, including a 10k warm-up phase to ensure training stability. 
In addition, we establish HIMOSA-light, which is a variant of HIMOSA that sets the number of experts to 4, significantly improving inference efficiency while ensuring model performance.

\begin{table*}[!ht]
\centering
\caption{
Quantitative results ($\times 4$) achieved by different methods on the AID datasets. 
Here, PSNR(dB)$\uparrow$ and SSIM$\uparrow$ values are reported. 
\textbf{bold} texts indicate the best performance.
}
\vspace{-2mm}
\resizebox{\textwidth}{!}{
    \begin{tabular}{lccccccccccccccc}
    \toprule
    \multirow{2}*{Category} & \multicolumn{2}{c}{SwinIR-light} & \multicolumn{2}{c}{ATD-light} & \multicolumn{2}{c}{HiT-SRF} & \multicolumn{2}{c}{CATANet} & \multicolumn{2}{c}{ESTNet} & \multicolumn{2}{c}{PFT-light}& \multicolumn{2}{c}{HIMOSA(Ours)}  \\
    \cmidrule(lr){2-3} \cmidrule(lr){4-5}\cmidrule(lr){6-7}\cmidrule(lr){8-9}\cmidrule(lr){10-11}\cmidrule(lr){12-13}\cmidrule(lr){14-15}
             & \makecell{PSNR} & \makecell{SSIM} & PSNR & \makecell{SSIM} & \makecell{PSNR} & SSIM & \makecell{PSNR}& \makecell{SSIM} & PSNR & \makecell{SSIM}&  \makecell{PSNR} & SSIM &  \makecell{PSNR} & SSIM \\
    \midrule
    Airport & 29.77 & 0.8165 & 29.91 & 0.8195 & 30.01 & 0.8232 & 30.01 & 0.8226 & 30.02 & 0.8229 & {30.06} & {0.8236} & \textbf{30.09} & \textbf{0.8252}\\
    Desert & 42.06 & 0.9517 & 42.10 & 0.9521 & 42.00 & 0.9521 & {42.12} & {0.9524} & 42.10 & 0.9521 & {42.13} & {0.9524} & \textbf{42.15} & \textbf{0.9525}\\
    Farmland & 36.29 & 0.8989 & 36.46 & 0.9009 & {36.60} & {0.9035} & 36.58 & 0.9030 & 36.56 & 0.9026 & {36.57} & {0.9027} & \textbf{36.68} & \textbf{0.9045}\\
    Forest & 29.61 & 0.7146 & 29.77 & 0.7211 & {29.80} & {0.7234} & {29.78} & 0.7232 & 29.77 & 0.7222 & {29.78} & {0.7224} & \textbf{29.81} & \textbf{0.7239}\\
    Industrial & 28.37 & 0.7866 & 28.65 & 0.7921 & 28.74 & 0.7980 & {28.76} & {0.7980} & 28.74 & 0.7972 & {28.79} & {0.7989} & \textbf{28.86} & \textbf{0.8021}\\
    Meadow & 36.60 & 0.8372 & 36.69 & 0.8381 & {36.72} & {0.8390} & 36.69 & 0.8385 & 36.69 & 0.8385 & {36.71} & {0.8386} & \textbf{36.74} & \textbf{0.8391}\\
    MediumResidential & 29.51 & 0.7680 & 29.79 & 0.7705 & {29.98} & {0.7825} & 29.92 & 0.7805 & 29.92 & 0.7809 & {29.91} & {0.7802} & \textbf{30.01} & \textbf{0.7834}\\
    Mountain & 30.69 & 0.7829 & 30.73 & 0.7833 & 30.76 & 0.7853 & 30.75 & 0.7849 & 30.75 & 0.7848 & {30.77} & {0.7856} & \textbf{30.79} & \textbf{0.7863}\\
    Park & 28.98 & 0.7623 & 29.11 & 0.7665 & {29.18} & {0.7706} & 29.18 & 0.7701 & 29.17 & 0.7699 & {29.19} & {0.7705} & \textbf{29.25} & \textbf{0.7728}\\
    Parking & 26.77 & 0.8327 & 27.81 & 0.8559 & {28.17} & {0.8634} & 28.01 & 0.8602 & 27.96 & 0.8587 & {28.07} & {0.8617} & \textbf{28.37} & \textbf{0.8679}\\
    Playground & 33.89 & 0.8761 & 34.15 & 0.8794 & {34.32} & {0.8854} & 34.27 & 0.8832 & 34.24 & 0.8838 & {34.31} & {0.8833} & \textbf{34.41} & \textbf{0.8867}\\
    Bareland & 38.48 & 0.8866 & 38.52 & 0.8867 & {38.52} & 0.8874 & 38.54 & {0.8876} & {38.56} & {0.8876} & {38.56} & {0.8876} & \textbf{38.58} & \textbf{0.8879}\\
    Pond & 31.58 & 0.8357 & 31.70 & 0.8374 & 31.72 & 0.8387 & {31.75} & {0.8389} & {31.77} & {0.8389} & {31.76} & {0.8392} & \textbf{31.78} & \textbf{0.8397}\\
    Port & 28.15 & 0.8442 & 28.41 & 0.8497 & {28.54} & {0.8543} & 28.51 & 0.8534 & 28.47 & 0.8525 & {28.56} & {0.8540} & \textbf{28.66} & \textbf{0.8569}\\
    RailwayStation & 28.56 & 0.7601 & 28.90 & 0.7680 & 28.93 & 0.7723 & {28.97} & 0.7724 & 28.93 & 0.7717 & {29.04} & 0.7740 & \textbf{29.06}  & \textbf{0.7758}\\
    Resort & 28.13 & 0.7712 & 28.28 & 0.7751 & {28.42} & {0.7808} & 28.39 & 0.7796 & 28.39 & 0.7798 & {28.41} & {0.7804} & \textbf{28.47} & \textbf{0.7829}\\
    River & 31.60 & 0.7891 & 31.68 & 0.7905 & {31.73} & {0.7926} & 31.72 & {0.7926} & 31.72 & 0.7923 & {31.73} & {0.7926} & \textbf{31.77} & \textbf{0.7936}\\
    School & 28.74 & 0.7962 & 28.95 & 0.8007 & 29.08 & {0.8061} & 29.05 & 0.8045 & 29.04 & 0.8049 & {29.07} & {0.8054} & \textbf{29.15} & \textbf{0.8082}\\
    SparseResidential & 27.26 & 0.6538 & 27.41 & 0.6592 & {27.49} & {0.6640} & 27.47 & 0.6633 & 27.48 & 0.6623 & {27.46} & {0.6626} & \textbf{27.51} & \textbf{0.6647}\\
    Square & 29.05 & 0.7988 & 29.39 & 0.8054 & 29.48 & 0.8097 & 29.45 & 0.8086 & 29.48 & 0.8091 & {29.54} & {0.8110} & \textbf{29.59} & \textbf{0.8125}\\
    Stadium & 28.44 & 0.8138 & 28.78 & 0.8211 & 28.74 & 0.8216 & 28.77 & 0.8222 & 28.79 & {0.8230} & \textbf{28.92} & \textbf{0.8270} & {28.89} & {0.8261}\\
    StorageTanks & 27.50 & 0.7664 & 27.64 & 0.7697 & {27.76} & {0.7750} & 27.73 & 0.7733 & 27.71 & 0.7740 & {27.75} & {0.7741} & \textbf{27.79} & \textbf{0.7763}\\
    BaseballField & 33.56 & 0.8903 & 33.78 & 0.8930 & {33.89} & {0.8952} & 33.88 & 0.8945 & {33.89} & 0.8945 & {33.91} & {0.8951} & \textbf{33.98} & \textbf{0.8961}\\
    Viaduct & 28.94 & 0.7533 & 29.11 & 0.7588 & {29.23} & {0.7652} & {29.23} & 0.7646 & {29.23} & {0.7652} & {29.27} & {0.7669} & \textbf{29.33} & \textbf{0.7693}\\
    Beach & 30.93 & 0.8063 & 31.04 & 0.8067 & 31.05 & {0.8079} & 31.06 & {0.8082} & 31.07 & 0.8079 & {31.07} & {0.8081} & \textbf{31.10} & \textbf{0.8087}\\
    Bridge & 31.80 & 0.8089 & 32.03 & 0.8112 & {32.08} & {0.8130} & 32.03 & 0.8119 & 32.07 & {0.8131} & {32.14} & {0.8140} & \textbf{32.15} & \textbf{0.8142}\\
    Center & 27.98 & 0.7928 & 28.31 & 0.8002 & {28.47} & {0.8064} & 28.42 & 0.8059 & 28.47 & 0.8069 & {28.48} & {0.8068} & \textbf{28.58} & \textbf{0.8101}\\
    Church & 25.32 & 0.7065 & 25.62 & 0.7186 & {25.78} & {0.7269} & 25.74 & 0.7246 & 25.75 & 0.7252 & {25.77} & {0.7254} & \textbf{25.84} & \textbf{0.7292}\\
    Commercial & 28.57 & 0.7944 & 28.78 & 0.7991 & {28.92} & {0.8050} & 28.89 & 0.8035 & 28.89 & 0.8039 & {28.90} & {0.8039} & \textbf{28.97} & \textbf{0.8067}\\
    DenseResidential & 25.20 & 0.7003 & 25.43 & 0.7117 & {25.59} & {0.7220} & 25.55 & 0.7185 & 25.52 & 0.7182 & {25.52} & {0.7178} & \textbf{25.64} & \textbf{0.7248}\\
    \midrule
    Average & 30.41 & 0.7999 & 30.63 & 0.8047 & {30.72} & {0.8090} & 30.71 & 0.8082 & 30.70 & 0.8082 & 30.74 & 0.8088 & \textbf{30.80} & \textbf{0.8109} \\
    \bottomrule
    \end{tabular}
}
\vspace{-4mm}
\label{tab:sota_category}
\end{table*}

\subsection{Results}
\noindent\textbf{Quantitative results.}
To validate the effectiveness of our method in remote sensing image super-resolution, we conduct comparisons with state-of-the-art methods in different remote sensing image datasets.
In \cref{tab:sota_category}, we compare our method with other natural image super-resolution methods SwinIR-light~\cite{liang2021swinir}, NLSA~\cite{nlsa}, ATD-light~\cite{ATDNet}, HiT-SRF~\cite{zhang2024hit}, PFT-light~\cite{PFTNet}, CATANet~\cite{CATANet}, and remote sensing image super-resolution method HSENet~\cite{hsenet}, TransENet~\cite{transenet}, ESTNet~\cite{ESTNet}. For a fair comparison, all models were retrained in the AID dataset. From \cref{tab:sota_category}, it can be seen that our method consistently achieves the best performance across various scene types, and outperforms the state-of-the-art methods by 0.06 dB and 0.0021 in average PSNR and SSIM, respectively. In particular, the results on scenes with highly redundant information, such as deserts and beaches, further confirm that not all tokens within a window need to be computed, verifying the effectiveness of the proposed sparse attention mechanism. In addition, the outstanding performance in dense residential areas and other scenes with evident multi-scale repetitive patterns demonstrates the capability of our method to effectively extract and leverage multi-scale information.

Furthermore, to validate the generalizability of our method, we conducted experiments on different remote sensing image datasets, including DOTA, NWPU, and DIOR. As shown in \cref{tab:sota_multidataset}, our method also achieves the best performance on these benchmarks.

\begin{table*}[!ht]
\centering
\caption{
Quantitative results achieved by different methods on the AID, DOTA, NWPU and DIOR datasets. 
Here, PSNR(dB)$\uparrow$, SSIM$\uparrow$ and LPIPS$\downarrow$ values are reported. 
\textbf{bold} and \underline{underline} texts indicate the best and the second-best performance, respectively.
}
\vspace{-2mm}
\resizebox{\textwidth}{!}{
    \begin{tabular}{lccccccccccccc}
    \toprule
    \multirow{2}*{Method} & \multirow{2}*{Scale} & \multicolumn{3}{c}{AID} & \multicolumn{3}{c}{DOTA} & \multicolumn{3}{c}{NWPU} & \multicolumn{3}{c}{DIOR}  \\
    \cmidrule(lr){3-5} \cmidrule(lr){6-8}\cmidrule(lr){9-11}\cmidrule(lr){12-14}
             &       & \makecell{PSNR} & \makecell{SSIM} & LPIPS & \makecell{PSNR}& \makecell{SSIM} & LPIPS & \makecell{PSNR}&  \makecell{SSIM} & LPIPS & \makecell{PSNR}&  \makecell{SSIM} & LPIPS\\
    \midrule
    \midrule
    NLSA (2021 CVPR)  &  $\times$2   & 36.74 & 0.9439 & 0.1087 & 39.75 & 0.9603 & 0.0881 & 34.71 & 0.9311 & 0.1189 & 37.41 & 0.9500 & 0.1083  \\
    SwinIR-light (2021 ICCVW) &  $\times$2   & 36.83 & 0.9428 & 0.1077 & 40.03 & 0.9598 & 0.0855 & 34.66 & 0.9229 & 0.1440 & 37.38 & 0.9490 & 0.1229 \\
    ATD-light (2024 CVPR) & $\times$2   & 36.97 & 0.9449 & 0.1101 & 40.10 & 0.9605 & 0.0879 & 34.83 & 0.9318 & 0.1129 & 37.59 & {0.9504} & 0.1099  \\
    ESTNet (2024 TIP) & $\times$2   & {37.00} & 0.9449 & 0.1071 & 40.14 & 0.9607 & 0.0854 & 34.82 & 0.9318 & 0.1110 & 37.60 & 0.9505 & 0.1075 \\
    HiT-SRF (2024 ECCV) & $\times$2   & 36.96 & {0.9456} & {0.1017} & 40.10 & 0.9605 & {0.0845} & 34.83 & 0.9318 & {0.1090} & 37.59 & {0.9504} & {0.1069} \\
    PFT-light (2025 CVPR) & $\times$2   & 36.90 & 0.9441 & 0.1075 & {40.19} & {0.9610} & 0.0855 & {34.84} & {0.9321} & 0.1427 & {37.63} &  {0.9508} & 0.1211  \\
    CATANet (2025 CVPR)  & $\times$2   & 36.97 & 0.9445 & 0.1093 & 40.07 & 0.9607 & 0.0878 & 34.83 & 0.9318 & 0.1126 & 37.60 & 0.9505 & 0.1104 \\
    \midrule
    HIMOSA-light (Ours) & $\times$2   & \underline{37.19} & \underline{0.9466} & \underline{0.1010} & \underline{40.22} & \underline{0.9610} & \underline{0.0808} &\underline{34.87} & \underline{0.9325} & \underline{0.1058} &\underline{37.69} & \underline{0.9512} & \underline{0.1057} \\
    HIMOSA (Ours) & $\times$2   & \textbf{37.20} & \textbf{0.9467} & \textbf{0.1005} & \textbf{40.26} & \textbf{0.9613} & \textbf{0.0788} &\textbf{34.90} & \textbf{0.9328} & \textbf{0.1043} &\textbf{37.70} &\textbf{0.9513} & \textbf{0.1044} \\
    \midrule
    \midrule
    HSENet (2021 TGRS)& $\times$4  & 29.51 & 0.7692 & 0.4024 & 31.31 & 0.8151 & 0.3482 & 28.31 & 0.7411 & 0.4265 & 29.46 & 0.7771 & 0.3977  \\
    TransENet (2021 TGRS)  &  $\times$4   & 29.53 & 0.7718 & 0.3528 & 31.56 & 0.8228 & 0.3036 & 28.29 & 0.7417 & 0.3767 & 29.37 & 0.7786 & 0.3509  \\
    NLSA (2021 CVPR)  &  $\times$4   & 30.55 & 0.8053 & \textbf{0.3100} & 32.92 & 0.8495 & {0.2681} & 29.23 & 0.7774 & 0.3254 & 30.54 & 0.8085 & 0.3223  \\
    SwinIR-light (2021 ICCVW) &  $\times$4   & 30.41 & 0.7999 & 0.3227 & 32.85 & 0.8465 & 0.2790 & 29.09 & 0.7716 & 0.3413 &30.37 &0.8041 & 0.3312  \\
    ATD-light (2024 CVPR) & $\times$4   & 30.63 & 0.8047 & 0.3231 & 33.17 & 0.8522 & 0.2755 &29.28 & 0.7785 & 0.3279 &30.67 &0.8088 & 0.3191  \\
    ESTNet (2024 TIP) & $\times$4   & 30.70 & 0.8082 & 0.3153 & 33.28 & 0.8548 & 0.2694 & 29.33 & 0.7807 & 0.3265 & 30.76 & 0.8119 & 0.3183   \\
    HiT-SRF (2024 ECCV) & $\times$4  & 30.72 & {0.8090} & 0.3129 & 33.29 & 0.8551 & 0.2689 & {29.36} & {0.7820} & \underline{0.3227} &30.74 & 0.8122 & \underline{0.3154}  \\
    PFT-light (2025 CVPR) & $\times$4   & {30.74} & 0.8088 & 0.3165 & {33.34} & {0.8555} & 0.2703 & 29.35 & 0.7815& 0.3364 & {30.77} & {0.8125} & 0.3252 \\
    CATANet (2025 CVPR)  & $\times$4   & 30.70 & 0.8082 & 0.3151 & 33.30 & 0.8549 & 0.2704 &29.33 & 0.7811& 0.3252 &30.75 &0.8119 & 0.3164 \\
    \midrule
    HIMOSA-light (Ours) & $\times$4   & \underline{30.78} & \underline{0.8103} & 0.3118 & \underline{33.37} & \underline{0.8563} & \underline{0.2673} &\underline{29.40} & \underline{0.7835} & {0.3269} &\underline{30.80} &\underline{0.8136} & 0.3214 \\
    HIMOSA (Ours) & $\times$4   & \textbf{30.80} & \textbf{0.8109} & \underline{0.3108} & \textbf{33.38} & \textbf{0.8567} & \textbf{0.2670} &\textbf{29.42} & \textbf{0.7840} & \textbf{0.3217} &\textbf{30.82} &\textbf{0.8141} & \textbf{0.3149} \\ 
    \bottomrule
    \end{tabular}
}
\vspace{-2mm}
\label{tab:sota_multidataset}
\end{table*}

\begin{figure*}[t]
\centering
\includegraphics[width=0.96  \textwidth]{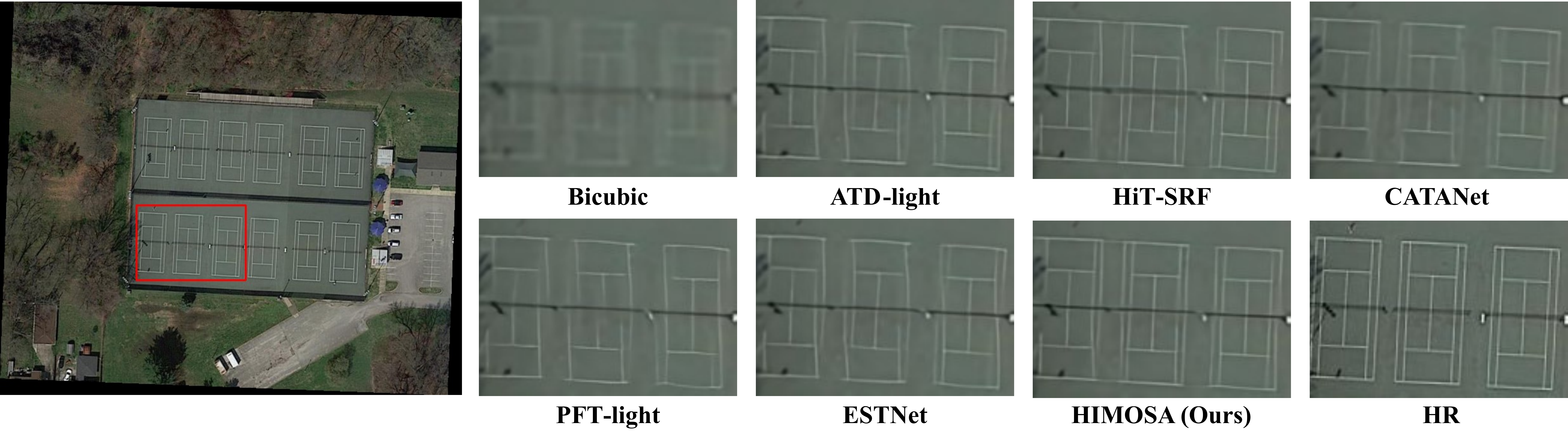}
\vspace{-4mm}
\caption{Visualization results ($\times 4$) achieved by different methods in DOTA datasets.}
\vspace{-4mm}
\label{fig.comparison_dota}
\end{figure*}


\noindent\textbf{Qualitative results.}
We present visual comparison results on the DOTA and AID datasets, as shown in \cref{fig.comparison_dota}, and \cref{fig.comparison_aid}. In the scenes with repetitive or weak-textured surfaces, our method achieves clearer and more accurate texture reconstruction compared to previous sparse Transformer-based approaches.
As illustrated in \cref{fig.comparison_dota}, the playground contains pronounced repetitive texture patterns, and our method achieves a more accurate reconstruction of surface textures compared to previous approaches, primarily because it suppresses redundant information within the window during inference and guides the network’s attention toward critical boundary details. Furthermore, by leveraging multi-scale windows, the network effectively exploits similar textures from other playground regions within the scene. In addition, as shown in \cref{fig.comparison_aid}, our method also produces clearer reconstruction results in weak-texture areas. These results provide strong evidence of the effectiveness of our design in handling repetitive and weak-texture scenes.

\subsection{Ablation Study}
In this subsection, we perform ablation studies on HIMOSA model and train all models for 250k iterations.

\noindent\textbf{Sparsity.} 
The sparsity is an important hyperparameter to balance computational complexity and model performance. When the sparsity approaches $1$, the attention becomes closer to dense attention, capturing more information but significantly increasing computational complexity. Conversely, as the sparsity increases, the model utilizes less information, thereby reducing computational cost but potentially degrading performance. Therefore, an appropriate choice of sparsity is essential to achieve an optimal trade-off between efficiency and reconstruction quality. 
\begin{table}[!t]
\begin{center}
\caption{Quantitative results ($\times 4$) achieved by methods with different sparsity.}
\vspace{-2mm}
\resizebox{\linewidth}{!}{
    \begin{tabular}{lccccc}
       \toprule
      \multirow{2}*{Sparsity} & \multicolumn{2}{c}{AID} & \multicolumn{2}{c}{DIOR} & \multirow{2}*{Inference time (ms)}\\
      \cmidrule(lr){2-3} \cmidrule(lr){4-5}
        & \makecell{PSNR} & \makecell{SSIM} & \makecell{PSNR} & \makecell{SSIM} \\
       \midrule
       $\rho_{(1,1,2,4,8,12)}$ & 30.80 & 0.8109 & 30.82 & 0.8141 & 93.30\\
       $\rho_{(1,1,2,4,8,8)}$ & 30.80 & 0.8110 & 30.83 & 0.8143 & 96.79\\
       $\rho_{(1,1,2,4,8,16)}$ & 30.80 & 0.8109 & 30.80 & 0.8136 & 92.58\\
       $\rho_{(1,2,4,8,12,16)}$ & 30.75 & 0.8094 & 30.78 & 0.8129 & 70.56\\
       
       \bottomrule
    \end{tabular}
}
\vspace{-8mm}
\label{table_sparsity}
\end{center}
\end{table}
We analyze the impact of different sparsity configurations on model performance and inference speed in \cref{table_sparsity}. After a comprehensive evaluation, we select the configuration (1, 1, 2, 4, 8, 12) as the sparsity setting for our method.

\noindent\textbf{Number of experts.}
We compared the impact of different numbers of experts. The results are shown in~\cref{table_experts}. Because more experts can provide a more refined distribution of image content, better performance is achieved. However, as the number of experts increases, the amount of token computation increases, resulting in a decrease in inference efficiency.
Therefore, in order to balance computational efficiency and model performance, we set the number of experts to 8. In addition, we extend HIMOSA to HIMOSA-light, adjusting the number of experts to 4.

\begin{table}[!t]
\begin{center}
\caption{Quantitative results ($\times 4$) achieved by methods with different numbers of experts.}
\vspace{-2mm}
\resizebox{\linewidth}{!}{
    \begin{tabular}{lccccc}
       \toprule
      \multirow{2}*{Num of experts} & \multicolumn{2}{c}{AID} & \multicolumn{2}{c}{DIOR}& \multirow{2}*{Inference time (ms)}\\
      \cmidrule(lr){2-3} \cmidrule(lr){4-5}
        & \makecell{PSNR} & \makecell{SSIM} & \makecell{PSNR}& \makecell{SSIM} &\\
       \midrule
       12 & 30.81 & 0.8113 & 30.84 & 0.8145 & 133.14\\
       8  & 30.80 & 0.8109 & 30.82 & 0.8141 & 93.30\\
       6 & 30.79 & 0.8107 & 30.81 & 0.8139 & 77.97\\
       4  & 30.78 & 0.8103 & 30.80 & 0.8136 & 63.82 \\
       \bottomrule
    \end{tabular}
}
\label{table_experts}
\vspace{-8mm}
\end{center}
\end{table}

\noindent\textbf{Content-aware routing.}
We analyze the content-aware routing mechanism. Specifically, we set different token selection strategies and analyze their impact on the performance of our method. We compare content-aware routing with two alternative selection strategies: random $k$ indices and the first $k$ indices in the token sequence. As shown in \cref{table_token_select}, the proposed content-aware routing achieves significant performance improvements.

In addition, we conducted a visualization analysis of the proposed content-aware routing mechanism. As shown in ~\cref{fig.sparse}, we visualized the token selection of the first HIMOSA block at the 4th, 5th, and 6th layers for the first four experts, where the red regions indicate the selected tokens. It can be observed that each expert adaptively selects tokens according to different image content. Furthermore, with the increase in sparsity, the method effectively avoids interference from redundant information within large windows while alleviating unnecessary computational cost.
\begin{table}[!t]
\begin{center}
\caption{Quantitative results ($\times 4$) achieved by methods with different token selection strategies.}
\vspace{-2mm}
\resizebox{\linewidth}{!}{
    \begin{tabular}{lccccccc}
       \toprule
      \multirow{2}*{Token selection strategy} & \multicolumn{2}{c}{AID} & \multicolumn{2}{c}{DIOR}\\
      \cmidrule(lr){2-3} \cmidrule(lr){4-5}
        & \makecell{PSNR} & \makecell{SSIM} & \makecell{PSNR}& \makecell{SSIM}\\
       \midrule
       content-aware routing selection & \textbf{30.80} & \textbf{0.8109} & \textbf{30.82} & \textbf{0.8141}\\
       random indices selection & 30.74 & 0.8091 & 30.78 & 0.8129 \\
       sequential indices selection & 30.73 & 0.8087 & 30.76 & 0.8121 \\
       \bottomrule
    \end{tabular}
}
\label{table_token_select}
\vspace{-6mm}
\end{center}
\end{table}

\begin{table}[!t]
\caption{Quantitative ablation study of different modules (PSNR $\uparrow$/SSIM $\uparrow$).}
\vspace{-2mm}
\resizebox{\linewidth}{!}
{
\centering
\setlength{\linewidth}{1.2pt}
\scriptsize
\begin{tabular}{c|cc|cc}
\toprule
Method & CA & ConvGLU & AID & DIOR \\
\midrule\midrule
\multirow{3}*{HIMOSA} & \textcolor{gray}{\Checkmark}  & \textcolor{gray}{\Checkmark} &  \textbf{30.80/0.8109} & \textbf{30.82/0.8141} \\
~ & \XSolidBrush & \textcolor{gray}{\Checkmark}  & 30.75/0.8094  & 30.80/0.8131 \\
~ & \textcolor{gray}{\Checkmark} & \XSolidBrush  & 30.78/0.8104  & 30.81/0.8137 \\
\bottomrule
\end{tabular}}
\label{tab:ablation}
\vspace{-2mm}
\end{table}

\begin{figure}[!t]
\centering
\includegraphics[width=0.96  \linewidth]{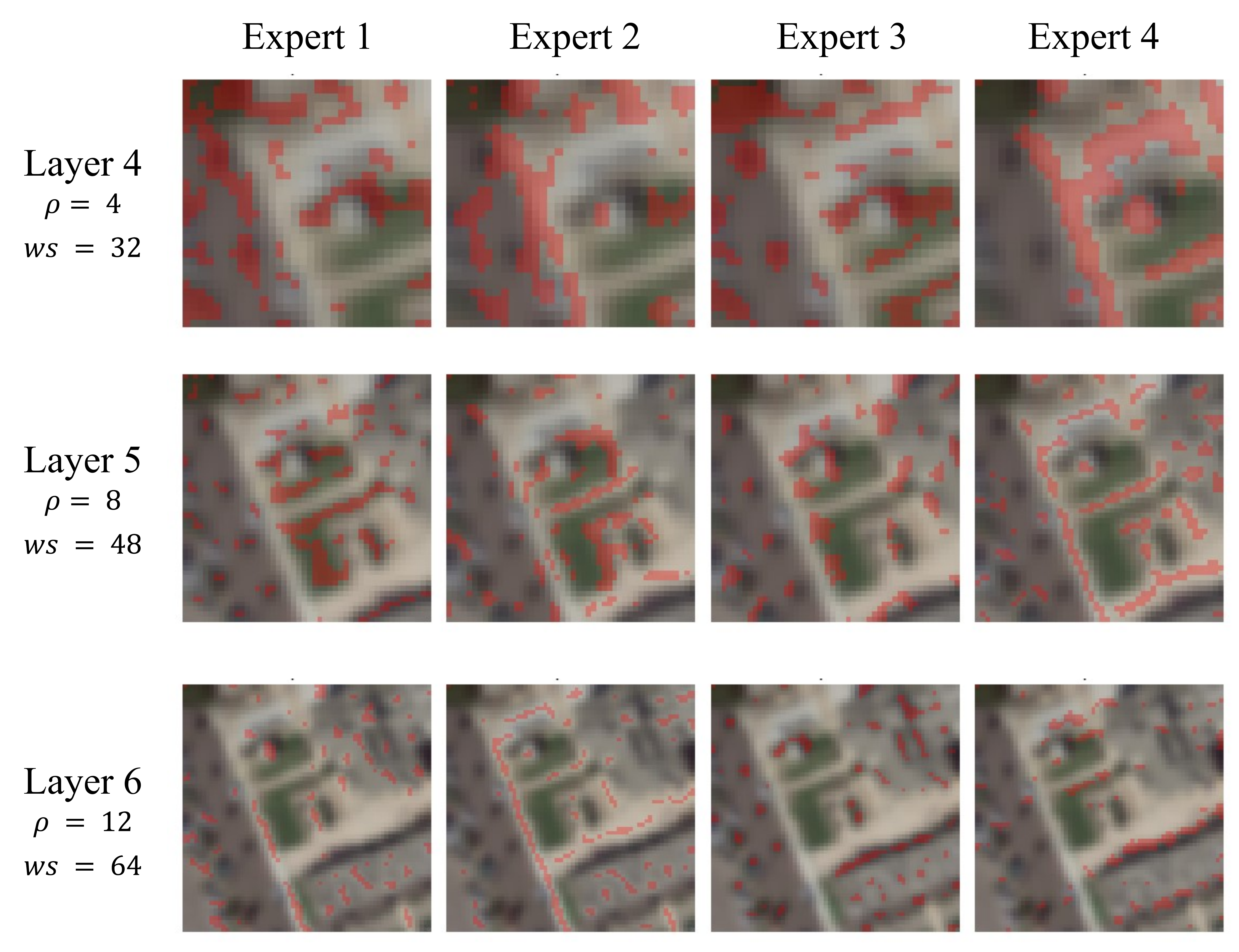}
\vspace{-4mm}
\caption{Visualization of sparse attention. The red region indicates the selected token.}
\vspace{-4mm}
\label{fig.sparse}
\end{figure}

\noindent\textbf{CA and ConvGLU.}
We perform ablation studies on the CA and ConvGLU modules. Specifically, we remove CA or replace ConvGLU with a simple feed-forward layer. As shown in \cref{tab:ablation}, both modules played a significant role in improving our model performance.

\subsection{Other results}
\noindent\textbf{Efficiency.}
To demonstrate the efficiency of our method, we compare our method with other lightweight methods in terms of parameters, FLOPs, and inference time. The evaluation was conducted using $256 \times 256$ image patches under the same experimental settings, and the results are presented in~\cref{table_effciency}. HIMOSA achieves comparable inference speed to other lightweight methods while significantly outperforming them in reconstruction performance. Moreover, although HIMOSA-light sacrifices a certain degree of performance, it achieves the fastest inference speed, exceeding the current best-performing method by 27.34\%.

\begin{figure}[t]
\centering
\includegraphics[width=0.96  \linewidth]{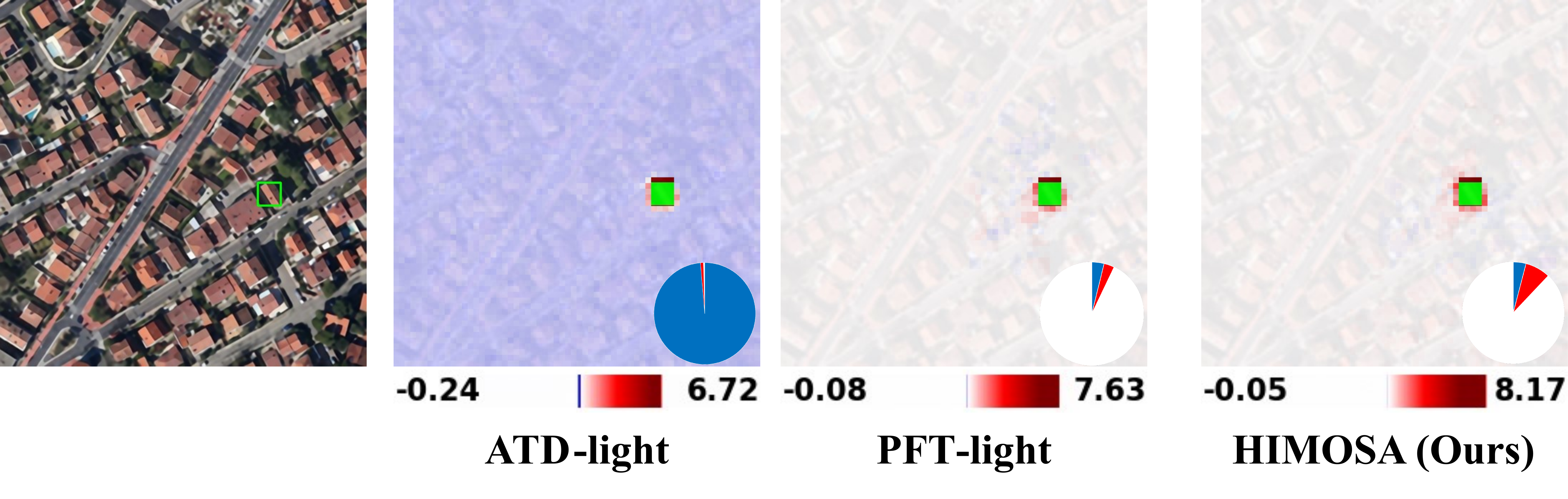}
\vspace{-2mm}
\caption{Causal effect maps (CEMs) of different methods on the AID \cite{xia2017aid} dataset. The patches with positive or negative causal effects and the ROI are indicated in red, blue, and green, respectively. The pie chart records the percentage of patches with different causal effects, and the colorbar expresses the effect range.}
\vspace{-6mm}
\label{fig.cem_results}
\end{figure}

\begin{table}[!t]
\begin{center}
\caption{Model parameters, FLOPs, and inference time of different methods ($\times 4$).}
\vspace{-2mm}
\resizebox{\linewidth}{!}{
    \begin{tabular}{lcccc}
       \toprule
       Methods & Params (M) & FLOPs (G) & Inference time (ms)\\
       \midrule
       SwinIR-light & 0.93 & 60.31 & 96.23\\
       ATD-light & 1.26 & 109.09 & 164.49\\
       HiT-SRF & 0.87 & 56.53 & 97.90\\
       CATANet & \textbf{0.53} & \textbf{46.77} & 87.84\\
       PFT-light & 1.10 & 98.32 & 371.90 \\
       ESTNet & 3.57 & 230.71 & 96.12 \\
       \midrule
       HIMOSA & 3.26 & 139.58 & 93.30\\
       HIMOSA-light & 2.14 & 66.71 & \textbf{63.82}\\
       \bottomrule
    \end{tabular}
}
\label{table_effciency}
\vspace{-4mm}
\end{center}
\end{table}
\noindent\textbf{Causal inference.}
In order to analyze the impact of redundant information on different networks, we used Causal Effect Map (CEM)~\cite{cem} to conduct attribution analysis on the networks. As illustrated in \cref{fig.cem_results}, while ATD-light can utilize a wider range of relevant information through its token dictionary, most of the information has been shown to have a negative impact. In contrast, our method effectively suppresses the participation of irrelevant or negative information and utilizes more beneficial information with hierarchical window design, resulting in superior performance.

\section{Conclusion}
In this work, we propose a lightweight framework for remote sensing image super-resolution. Specifically, HIMOSA exploits the inherent redundancy of remote sensing imagery by introducing a content-aware sparse attention mechanism, achieving a lightweight design while preserving strong reconstruction capability. To further tackle the challenge of multi-scale repetitive patterns, HIMOSA incorporates a progressive window expansion strategy, where the sparsity of the attention mechanism is adaptively adjusted to reduce computational cost. Moreover, we extend our framework to HIMOSA-light, which attains significant improvements in inference efficiency with minimal performance degradation. Extensive experiments conducted on multiple remote sensing datasets demonstrate that our method achieves state-of-the-art performance while maintaining superior computational efficiency.

{
    \small
    \bibliographystyle{ieeenat_fullname}
    \bibliography{main}

@String(CVPR= {IEEE Conf. Comput. Vis. Pattern Recog.})

@String(ECCV= {Eur. Conf. Comput. Vis.})

@String(ICLR = {Int. Conf. Learn. Represent.})

@String(CVPR  = {CVPR})

@String(ECCV  = {ECCV})

@String(ICLR  = {ICLR})

@ARTICLE{TTST,
  author={Xiao, Yi and Yuan, Qiangqiang and Jiang, Kui and He, Jiang and Lin, Chia-Wen and Zhang, Liangpei},
  journal={IEEE Transactions on Image Processing}, 
  title={TTST: A Top-k Token Selective Transformer for Remote Sensing Image Super-Resolution}, 
  year={2024},
  volume={33},
  number={},
  pages={738-752},
  doi={10.1109/TIP.2023.3349004}
}

@InProceedings{chen2023activating,
    author = {Chen, Xiangyu and Wang, Xintao and Zhou, Jiantao and Qiao, Yu and Dong, Chao},
    title = {Activating More Pixels in Image Super-Resolution Transformer},
    booktitle = {Proceedings of the IEEE/CVF Conference on Computer Vision and Pattern Recognition (CVPR)},
    month = {June},
    year = {2023},
    pages = {22367-22377}
}

@inproceedings{liang2021swinir,
  title={Swinir: Image restoration using swin transformer},
  author={Liang, Jingyun and Cao, Jiezhang and Sun, Guolei and Zhang, Kai and Van Gool, Luc and Timofte, Radu},
  booktitle={Proceedings of the IEEE/CVF International Conference on Computer Vision},
  pages={1833--1844},
  year={2021}
}

@article{zhangSwinFIRRevisitingSwinIR2023a,
  title={Swinfir: Revisiting the swinir with fast fourier convolution and improved training for image super-resolution},
  author={Zhang, Dafeng and Huang, Feiyu and Liu, Shizhuo and Wang, Xiaobing and Jin, Zhezhu},
  journal={arXiv preprint arXiv:2208.11247},
  year={2022}
}

@inproceedings{zhang2024hit,
  title={HiT-SR: Hierarchical transformer for efficient image super-resolution},
  author={Zhang, Xiang and Zhang, Yulun and Yu, Fisher},
  booktitle={European Conference on Computer Vision},
  pages={483--500},
  year={2024},
  organization={Springer}
}

@InProceedings{CATANet,
  author = {Liu, Xin and Liu, Jie and Tang, Jie and Wu, Gangshan},
  title = {CATANet: Efficient Content-Aware Token Aggregation for Lightweight Image Super-Resolution},
  booktitle = {Proceedings of the IEEE/CVF Conference on Computer Vision and Pattern Recognition (CVPR)},
  month = {June},
  year = {2025},
  pages = {17902-17912}
}

@ARTICLE{ESTNet,
  author={Kang, Xudong and Duan, Puhong and Li, Jier and Li, Shutao},
  journal={IEEE Transactions on Image Processing}, 
  title={Efficient Swin Transformer for Remote Sensing Image Super-Resolution}, 
  year={2024},
  volume={33},
  number={},
  pages={6367-6379},
  doi={10.1109/TIP.2024.3489228}}

@INPROCEEDINGS{ATDNet,
  author={Zhang, Leheng and Li, Yawei and Zhou, Xingyu and Zhao, Xiaorui and Gu, Shuhang},
  booktitle={2024 IEEE/CVF Conference on Computer Vision and Pattern Recognition (CVPR)}, 
  title={Transcending the Limit of Local Window: Advanced Super-Resolution Transformer with Adaptive Token Dictionary}, 
  year={2024},
  volume={},
  number={},
  pages={2856-2865},
  doi={10.1109/CVPR52733.2024.00276}}

@INPROCEEDINGS{PFTNet,
  author={Long, Wei and Zhou, Xingyu and Zhang, Leheng and Gu, Shuhang},
  booktitle={2025 IEEE/CVF Conference on Computer Vision and Pattern Recognition (CVPR)}, 
  title={Progressive Focused Transformer for Single Image Super-Resolution}, 
  year={2025},
  volume={},
  number={},
  pages={2279-2288},
  doi={10.1109/CVPR52734.2025.00218}}

@inproceedings{art,
  title={Accurate Image Restoration with Attention Retractable Transformer},
  author={Zhang, Jiale and Zhang, Yulun and Gu, Jinjin and Zhang, Yongbing and Kong, Linghe and Yuan, Xin},
  booktitle={ICLR},
  year={2023}
}

@article{srcnn,
  title={Image super-resolution using deep convolutional networks},
  author={Dong, Chao and Loy, Chen Change and He, Kaiming and Tang, Xiaoou},
  journal={IEEE transactions on pattern analysis and machine intelligence},
  volume={38},
  number={2},
  pages={295--307},
  year={2015},
  publisher={IEEE}
}

@inproceedings{nlsa,
  title={Image super-resolution with non-local sparse attention},
  author={Mei, Yiqun and Fan, Yuchen and Zhou, Yuqian},
  booktitle={Proceedings of the IEEE/CVF conference on computer vision and pattern recognition},
  pages={3517--3526},
  year={2021}
}

@inproceedings{DRSformer,
  title={Learning a sparse transformer network for effective image deraining},
  author={Chen, Xiang and Li, Hao and Li, Mingqiang and Pan, Jinshan},
  booktitle={Proceedings of the IEEE/CVF conference on computer vision and pattern recognition},
  pages={5896--5905},
  year={2023}
}

@article{zhou2022mixture,
  title={Mixture-of-experts with expert choice routing},
  author={Zhou, Yanqi and Lei, Tao and Liu, Hanxiao and Du, Nan and Huang, Yanping and Zhao, Vincent and Dai, Andrew M and Le, Quoc V and Laudon, James and others},
  journal={Advances in Neural Information Processing Systems},
  volume={35},
  pages={7103--7114},
  year={2022}
}

@article{csordas2024switchhead,
  title={Switchhead: Accelerating transformers with mixture-of-experts attention},
  author={Csord{\'a}s, R{\'o}bert and Pi{\k{e}}kos, Piotr and Irie, Kazuki and Schmidhuber, J{\"u}rgen},
  journal={Advances in Neural Information Processing Systems},
  volume={37},
  pages={74411--74438},
  year={2024}
}

@article{fu2024moa,
  title={Moa: Mixture of sparse attention for automatic large language model compression},
  author={Fu, Tianyu and Huang, Haofeng and Ning, Xuefei and Zhang, Genghan and Chen, Boju and Wu, Tianqi and Wang, Hongyi and Huang, Zixiao and Li, Shiyao and Yan, Shengen and others},
  journal={arXiv preprint arXiv:2406.14909},
  year={2024}
}

@article{pikekos2025mosa,
  title={Mixture of Sparse Attention: Content-Based Learnable Sparse Attention via Expert-Choice Routing},
  author={Pi{\k{e}}kos, Piotr and Csord{\'a}s, R{\'o}bert and Schmidhuber, J{\"u}rgen},
  journal={arXiv preprint arXiv:2505.00315},
  year={2025}
}

@inproceedings{wang2022kvt,
  title={Kvt: k-nn attention for boosting vision transformers},
  author={Wang, Pichao and Wang, Xue and Wang, Fan and Lin, Ming and Chang, Shuning and Li, Hao and Jin, Rong},
  booktitle={European conference on computer vision},
  pages={285--302},
  year={2022},
  organization={Springer}
}

@article{zhao2019explicit,
  title={Explicit sparse transformer: Concentrated attention through explicit selection},
  author={Zhao, Guangxiang and Lin, Junyang and Zhang, Zhiyuan and Ren, Xuancheng and Su, Qi and Sun, Xu},
  journal={arXiv preprint arXiv:1912.11637},
  year={2019}
}

@article{ainslie2023gqa,
  title={Gqa: Training generalized multi-query transformer models from multi-head checkpoints},
  author={Ainslie, Joshua and Lee-Thorp, James and De Jong, Michiel and Zemlyanskiy, Yury and Lebr{\'o}n, Federico and Sanghai, Sumit},
  journal={arXiv preprint arXiv:2305.13245},
  year={2023}
}

@inproceedings{zamfir2024seemore,
  title={See More Details: Efficient Image Super-Resolution by Experts Mining}, 
  author={Eduard Zamfir and Zongwei Wu and Nancy Mehta and Yulun Zhang and Radu Timofte},
  booktitle={International Conference on Machine Learning},
  year={2024},
  organization={PMLR}
}

@inproceedings{fsrcnn,
  title={Accelerating the super-resolution convolutional neural network},
  author={Dong, Chao and Loy, Chen Change and Tang, Xiaoou},
  booktitle={European conference on computer vision},
  pages={391--407},
  year={2016},
  organization={Springer}
}

@inproceedings{espcn,
  title={Real-time single image and video super-resolution using an efficient sub-pixel convolutional neural network},
  author={Shi, Wenzhe and Caballero, Jose and Husz{\'a}r, Ferenc and Totz, Johannes and Aitken, Andrew P and Bishop, Rob and Rueckert, Daniel and Wang, Zehan},
  booktitle={Proceedings of the IEEE conference on computer vision and pattern recognition},
  pages={1874--1883},
  year={2016}
}

@inproceedings{vdsr,
  title={Accurate image super-resolution using very deep convolutional networks},
  author={Kim, Jiwon and Lee, Jung Kwon and Lee, Kyoung Mu},
  booktitle={Proceedings of the IEEE conference on computer vision and pattern recognition},
  pages={1646--1654},
  year={2016}
}

@inproceedings{drcn,
  title={Deeply-recursive convolutional network for image super-resolution},
  author={Kim, Jiwon and Lee, Jung Kwon and Lee, Kyoung Mu},
  booktitle={Proceedings of the IEEE conference on computer vision and pattern recognition},
  pages={1637--1645},
  year={2016}
}

@inproceedings{drrn,
  title={Image super-resolution via deep recursive residual network},
  author={Tai, Ying and Yang, Jian and Liu, Xiaoming},
  booktitle={Proceedings of the IEEE conference on computer vision and pattern recognition},
  pages={3147--3155},
  year={2017}
}

@inproceedings{srdensenet,
  title={Image super-resolution using dense skip connections},
  author={Tong, Tong and Li, Gen and Liu, Xiejie and Gao, Qinquan},
  booktitle={Proceedings of the IEEE international conference on computer vision},
  pages={4799--4807},
  year={2017}
}

@inproceedings{srresnet,
  title={Photo-realistic single image super-resolution using a generative adversarial network},
  author={Ledig, Christian and Theis, Lucas and Husz{\'a}r, Ferenc and Caballero, Jose and Cunningham, Andrew and Acosta, Alejandro and Aitken, Andrew and Tejani, Alykhan and Totz, Johannes and Wang, Zehan and others},
  booktitle={Proceedings of the IEEE conference on computer vision and pattern recognition},
  pages={4681--4690},
  year={2017}
}

@inproceedings{edsr,
  title={Enhanced deep residual networks for single image super-resolution},
  author={Lim, Bee and Son, Sanghyun and Kim, Heewon and Nah, Seungjun and Mu Lee, Kyoung},
  booktitle={Proceedings of the IEEE conference on computer vision and pattern recognition workshops},
  pages={136--144},
  year={2017}
}

@inproceedings{rdn,
  title={Residual dense network for image super-resolution},
  author={Zhang, Yulun and Tian, Yapeng and Kong, Yu and Zhong, Bineng and Fu, Yun},
  booktitle={Proceedings of the IEEE conference on computer vision and pattern recognition},
  pages={2472--2481},
  year={2018}
}

@inproceedings{rcan,
  title={Image super-resolution using very deep residual channel attention networks},
  author={Zhang, Yulun and Li, Kunpeng and Li, Kai and Wang, Lichen and Zhong, Bineng and Fu, Yun},
  booktitle={Proceedings of the European conference on computer vision (ECCV)},
  pages={286--301},
  year={2018}
}

@inproceedings{esrgan,
  title={Esrgan: Enhanced super-resolution generative adversarial networks},
  author={Wang, Xintao and Yu, Ke and Wu, Shixiang and Gu, Jinjin and Liu, Yihao and Dong, Chao and Qiao, Yu and Change Loy, Chen},
  booktitle={Proceedings of the European conference on computer vision (ECCV) workshops},
  pages={0--0},
  year={2018}
}

@inproceedings{holistic-attenion,
  title={Single image super-resolution via a holistic attention network},
  author={Niu, Ben and Wen, Weilei and Ren, Wenqi and Zhang, Xiangde and Yang, Lianping and Wang, Shuzhen and Zhang, Kaihao and Cao, Xiaochun and Shen, Haifeng},
  booktitle={European conference on computer vision},
  pages={191--207},
  year={2020},
  organization={Springer}
}

@article{hsenet,
  title={Hybrid-scale self-similarity exploitation for remote sensing image super-resolution},
  author={Lei, Sen and Shi, Zhenwei},
  journal={IEEE Transactions on Geoscience and Remote Sensing},
  volume={60},
  pages={1--10},
  year={2021},
  publisher={IEEE}
}

@inproceedings{swintransformer,
  title={Swin transformer: Hierarchical vision transformer using shifted windows},
  author={Liu, Ze and Lin, Yutong and Cao, Yue and Hu, Han and Wei, Yixuan and Zhang, Zheng and Lin, Stephen and Guo, Baining},
  booktitle={Proceedings of the IEEE/CVF international conference on computer vision},
  pages={10012--10022},
  year={2021}
}

@article{vit,
  title={An image is worth 16x16 words: Transformers for image recognition at scale},
  author={Dosovitskiy, Alexey and Beyer, Lucas and Kolesnikov, Alexander and Weissenborn, Dirk and Zhai, Xiaohua and Unterthiner, Thomas and Dehghani, Mostafa and Minderer, Matthias and Heigold, Georg and Gelly, Sylvain and others},
  journal={arXiv preprint arXiv:2010.11929},
  year={2020}
}

@inproceedings{ipt,
  title={Pre-trained image processing transformer},
  author={Chen, Hanting and Wang, Yunhe and Guo, Tianyu and Xu, Chang and Deng, Yiping and Liu, Zhenhua and Ma, Siwei and Xu, Chunjing and Xu, Chao and Gao, Wen},
  booktitle={Proceedings of the IEEE/CVF conference on computer vision and pattern recognition},
  pages={12299--12310},
  year={2021}
}

@inproceedings{restormer,
  title={Restormer: Efficient transformer for high-resolution image restoration},
  author={Zamir, Syed Waqas and Arora, Aditya and Khan, Salman and Hayat, Munawar and Khan, Fahad Shahbaz and Yang, Ming-Hsuan},
  booktitle={Proceedings of the IEEE/CVF conference on computer vision and pattern recognition},
  pages={5728--5739},
  year={2022}
}

@inproceedings{yoo2023enriched,
  title={Enriched cnn-transformer feature aggregation networks for super-resolution},
  author={Yoo, Jinsu and Kim, Taehoon and Lee, Sihaeng and Kim, Seung Hwan and Lee, Honglak and Kim, Tae Hyun},
  booktitle={Proceedings of the IEEE/CVF winter conference on applications of computer vision},
  pages={4956--4965},
  year={2023}
}

@article{cat,
  title={Cross aggregation transformer for image restoration},
  author={Chen, Zheng and Zhang, Yulun and Gu, Jinjin and Kong, Linghe and Yuan, Xin and others},
  journal={Advances in Neural Information Processing Systems},
  volume={35},
  pages={25478--25490},
  year={2022}
}

@article{transenet,
  title={Transformer-based multistage enhancement for remote sensing image super-resolution},
  author={Lei, Sen and Shi, Zhenwei and Mo, Wenjing},
  journal={IEEE Transactions on Geoscience and Remote Sensing},
  volume={60},
  pages={1--11},
  year={2021},
  publisher={IEEE}
}

@article{shazeer2017outrageously,
  title={Outrageously large neural networks: The sparsely-gated mixture-of-experts layer},
  author={Shazeer, Noam and Mirhoseini, Azalia and Maziarz, Krzysztof and Davis, Andy and Le, Quoc and Hinton, Geoffrey and Dean, Jeff},
  journal={arXiv preprint arXiv:1701.06538},
  year={2017}
}

@article{fedus2022switch,
  title={Switch transformers: Scaling to trillion parameter models with simple and efficient sparsity},
  author={Fedus, William and Zoph, Barret and Shazeer, Noam},
  journal={Journal of Machine Learning Research},
  volume={23},
  number={120},
  pages={1--39},
  year={2022}
}

@article{lepikhin2020gshard,
  title={Gshard: Scaling giant models with conditional computation and automatic sharding},
  author={Lepikhin, Dmitry and Lee, HyoukJoong and Xu, Yuanzhong and Chen, Dehao and Firat, Orhan and Huang, Yanping and Krikun, Maxim and Shazeer, Noam and Chen, Zhifeng},
  journal={arXiv preprint arXiv:2006.16668},
  year={2020}
}

@article{xia2017aid,
  title={AID: A benchmark data set for performance evaluation of aerial scene classification},
  author={Xia, Gui-Song and Hu, Jingwen and Hu, Fan and Shi, Baoguang and Bai, Xiang and Zhong, Yanfei and Zhang, Liangpei and Lu, Xiaoqiang},
  journal={IEEE Transactions on Geoscience and Remote Sensing},
  volume={55},
  number={7},
  pages={3965--3981},
  year={2017},
  publisher={IEEE}
}

@inproceedings{xia2018dota,
  title={DOTA: A large-scale dataset for object detection in aerial images},
  author={Xia, Gui-Song and Bai, Xiang and Ding, Jian and Zhu, Zhen and Belongie, Serge and Luo, Jiebo and Datcu, Mihai and Pelillo, Marcello and Zhang, Liangpei},
  booktitle={Proceedings of the IEEE conference on computer vision and pattern recognition},
  pages={3974--3983},
  year={2018}
}

@article{li2020object,
  title={Object detection in optical remote sensing images: A survey and a new benchmark},
  author={Li, Ke and Wan, Gang and Cheng, Gong and Meng, Liqiu and Han, Junwei},
  journal={ISPRS journal of photogrammetry and remote sensing},
  volume={159},
  pages={296--307},
  year={2020},
  publisher={Elsevier}
}

@article{cheng2017remote,
  title={Remote sensing image scene classification: Benchmark and state of the art},
  author={Cheng, Gong and Han, Junwei and Lu, Xiaoqiang},
  journal={Proceedings of the IEEE},
  volume={105},
  number={10},
  pages={1865--1883},
  year={2017},
  publisher={IEEE}
}

@inproceedings{lam,
  title={Interpreting Super-Resolution Networks with Local Attribution Maps},
  author={Gu, Jinjin and Dong, Chao},
  booktitle={Proceedings of the IEEE/CVF Conference on Computer Vision and Pattern Recognition},
  pages={9199--9208},
  year={2021}
}

@article{cem,
  title={Interpreting low-level vision models with causal effect maps},
  author={Hu, Jinfan and Gu, Jinjin and Yu, Shiyao and Yu, Fanghua and Li, Zheyuan and You, Zhiyuan and Lu, Chaochao and Dong, Chao},
  journal={IEEE Transactions on Pattern Analysis and Machine Intelligence},
  year={2025},
  publisher={IEEE}
}

@article{muon,
  title={Muon is scalable for LLM training},
  author={Liu, Jingyuan and Su, Jianlin and Yao, Xingcheng and Jiang, Zhejun and Lai, Guokun and Du, Yulun and Qin, Yidao and Xu, Weixin and Lu, Enzhe and Yan, Junjie and others},
  journal={arXiv preprint arXiv:2502.16982},
  year={2025}
}

@article{moe,
  title={Adaptive mixtures of local experts},
  author={Jacobs, Robert A and Jordan, Michael I and Nowlan, Steven J and Hinton, Geoffrey E},
  journal={Neural computation},
  volume={3},
  number={1},
  pages={79--87},
  year={1991},
  publisher={MIT Press}
}

@inproceedings{pixelshuffle,
  title={Real-time single image and video super-resolution using an efficient sub-pixel convolutional neural network},
  author={Shi, Wenzhe and Caballero, Jose and Husz{\'a}r, Ferenc and Totz, Johannes and Aitken, Andrew P and Bishop, Rob and Rueckert, Daniel and Wang, Zehan},
  booktitle={Proceedings of the IEEE conference on computer vision and pattern recognition},
  pages={1874--1883},
  year={2016}
}

@article{transformer,
  title={Attention is all you need},
  author={Vaswani, Ashish and Shazeer, Noam and Parmar, Niki and Uszkoreit, Jakob and Jones, Llion and Gomez, Aidan N and Kaiser, {\L}ukasz and Polosukhin, Illia},
  journal={Advances in neural information processing systems},
  volume={30},
  year={2017}
}

@inproceedings{shi2024transnext,
  title={Transnext: Robust foveal visual perception for vision transformers},
  author={Shi, Dai},
  booktitle={Proceedings of the IEEE/CVF conference on computer vision and pattern recognition},
  pages={17773--17783},
  year={2024}
}

@inproceedings{stereoinr,
  title={Stereoinr: Cross-view geometry consistent stereo super resolution with implicit neural representation},
  author={Liu, Yi and Liu, Xinyi and Wan, Yi and Xia, Panwang and Wu, Qiong and Zhang, Yongjun},
  booktitle={Proceedings of the 33rd ACM International Conference on Multimedia},
  pages={1003--1012},
  year={2025}
}
}

\clearpage
\setcounter{page}{1}
\maketitlesupplementary
\appendix

\section{Training details}
\label{sec:datasets}
The details of the datasets used in our experiments are summarized in \cref{table_datasets}. Specifically, we evaluate the proposed HIMOSA framework on several representative remote sensing image super-resolution benchmarks to ensure the generalization and robustness of our method. These datasets cover diverse scenes, including urban areas, agricultural regions, coastal zones, and mountainous terrains, thereby providing a comprehensive assessment under various spatial structures and texture complexities. Each dataset contains high-resolution (HR) images and their corresponding low-resolution (LR) counterparts generated through bicubic downsampling with different scale factors. During training, we adopt random cropping, rotation, and horizontal flipping for data augmentation to improve the model’s robustness and generalization ability.

We adopt a multi-step learning rate decay strategy, initializing the learning rate at $5\times10^{-4}$ and reducing it by a factor of 0.5 at iterations 150K, 200K, 225K, and 240K. We set the input patch size to $64\times64$ and use random rotation and horizontally flipping for data augmentation. The minibatch size is set to 16. During training, we employ the L1 loss function to optimize the network. All experiments are conducted on four NVIDIA RTX 4090 GPUs.


\section{Local attribute analysis}
\label{sec:lam}
\begin{figure}[!t]
\centering
\includegraphics[width=0.96  \linewidth]{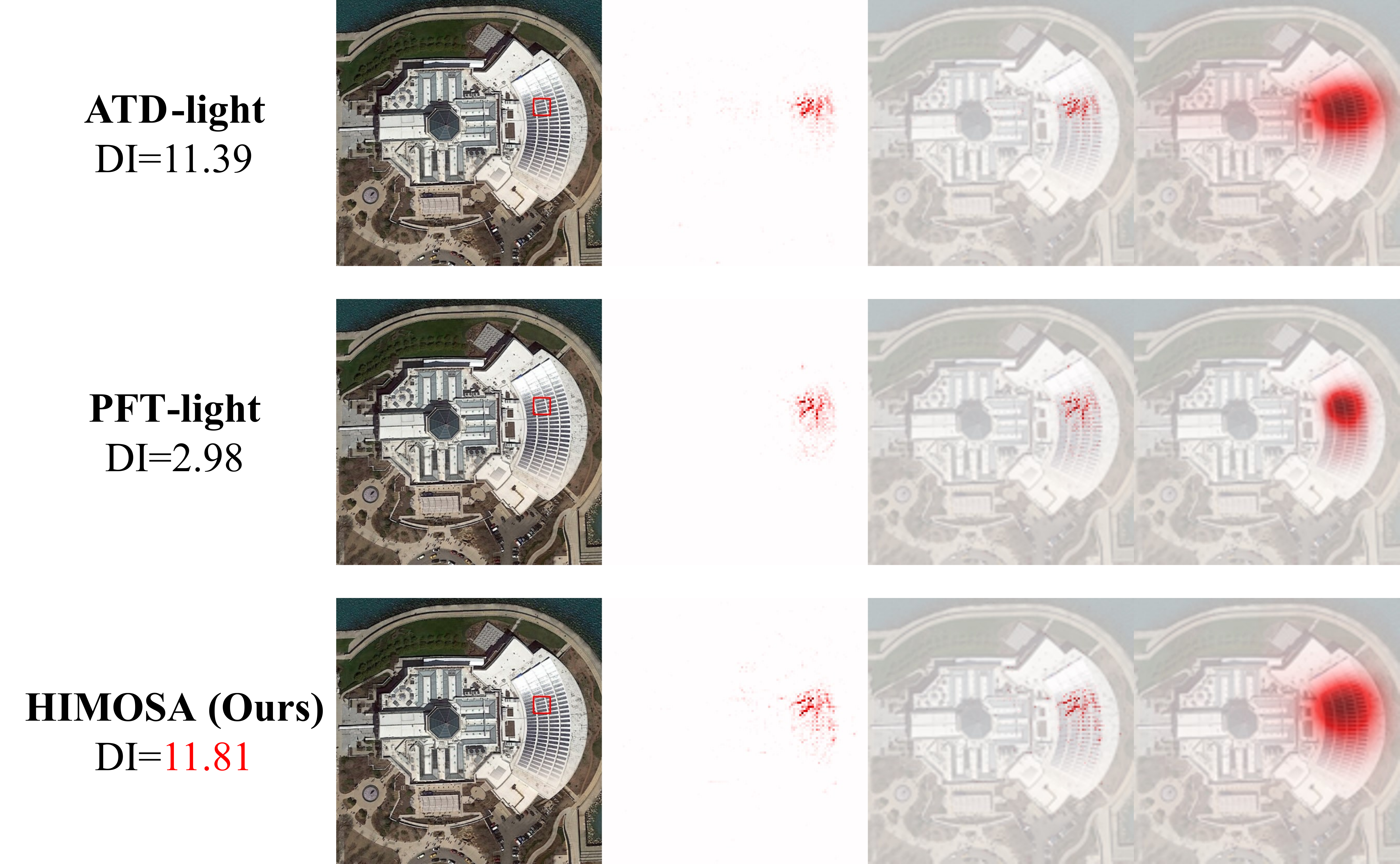}
\vspace{-2mm}
\caption{Local attribution maps (LAMs) of different methods on the AID \cite{xia2017aid} dataset. DI (Diffusion Index) indicates the range of involved pixels, with a higher DI indicating a wider receptive field.}
\vspace{-2mm}
\label{fig.lam_results}
\end{figure}

To further interpret the internal behavior of our model, we employ the Local Attribution Map (LAM)~\cite{lam}, an explainability tool designed for super-resolution tasks. LAM identifies which pixels in the low-resolution (LR) input contribute most to the generation of the high-resolution (HR) output, thereby revealing how effectively a model utilizes spatial information. As shown in ~\cref{fig.lam_results}, our HIMOSA exhibits a broader and denser distribution of informative regions, indicating that it can capture a larger set of relevant pixels while suppressing noise and redundant information. Compared to previous methods, HIMOSA demonstrates a higher Diffusion Index (DI), reflecting its superior ability to focus on meaningful contextual cues. These observations confirm that our content-aware sparse attention not only improves quantitative metrics but also enhances the interpretability and reliability of the reconstruction process.

\begin{algorithm}[!t]
 \caption{HIMOSA Block}
 \begin{algorithmic}[1]
 \renewcommand{\algorithmicrequire}{\textbf{Input:}}
 \renewcommand{\algorithmicensure}{\textbf{Output:}}
 \REQUIRE $\boldsymbol{X}$ input feature; $\boldsymbol{\mathit{ws}_{B}}$ base window size; $(\boldsymbol{\alpha_{0},\alpha_{1},..., \alpha_{M}})$ hierarchical ratio; $\boldsymbol{M}$ num of hierarchical layers.
 \ENSURE  $\boldsymbol{X}_{M}$ output feature;
 \FOR {$i = 0$ to $M$}
 \STATE Channel Attention: $X_{\mathrm{CAB}} = \mathrm{CAB}(X_{i})$;
 \STATE Partition windows with $\mathit{ws}_{i} = \alpha_{i}\mathit{ws}_{B}$;
 \STATE Content-aware routing sparse attention in window $\mathit{ws}_{i}$: $X_\mathrm{CARSA} = \mathrm{CARSA}(X_{i},A_{i})$;
 \STATE $X_{i} = X_{i}+X_\mathrm{CARSA}+X_\mathrm{CAB}$;
 \STATE $X_{i+1} = \mathrm{ConvGLU}(X_{i})+X_{i}$;
 \ENDFOR
 \end{algorithmic} 
 \end{algorithm}

\section{Other visualization results}
\begin{figure*}[t]
\centering
\includegraphics[width=0.96  \textwidth]{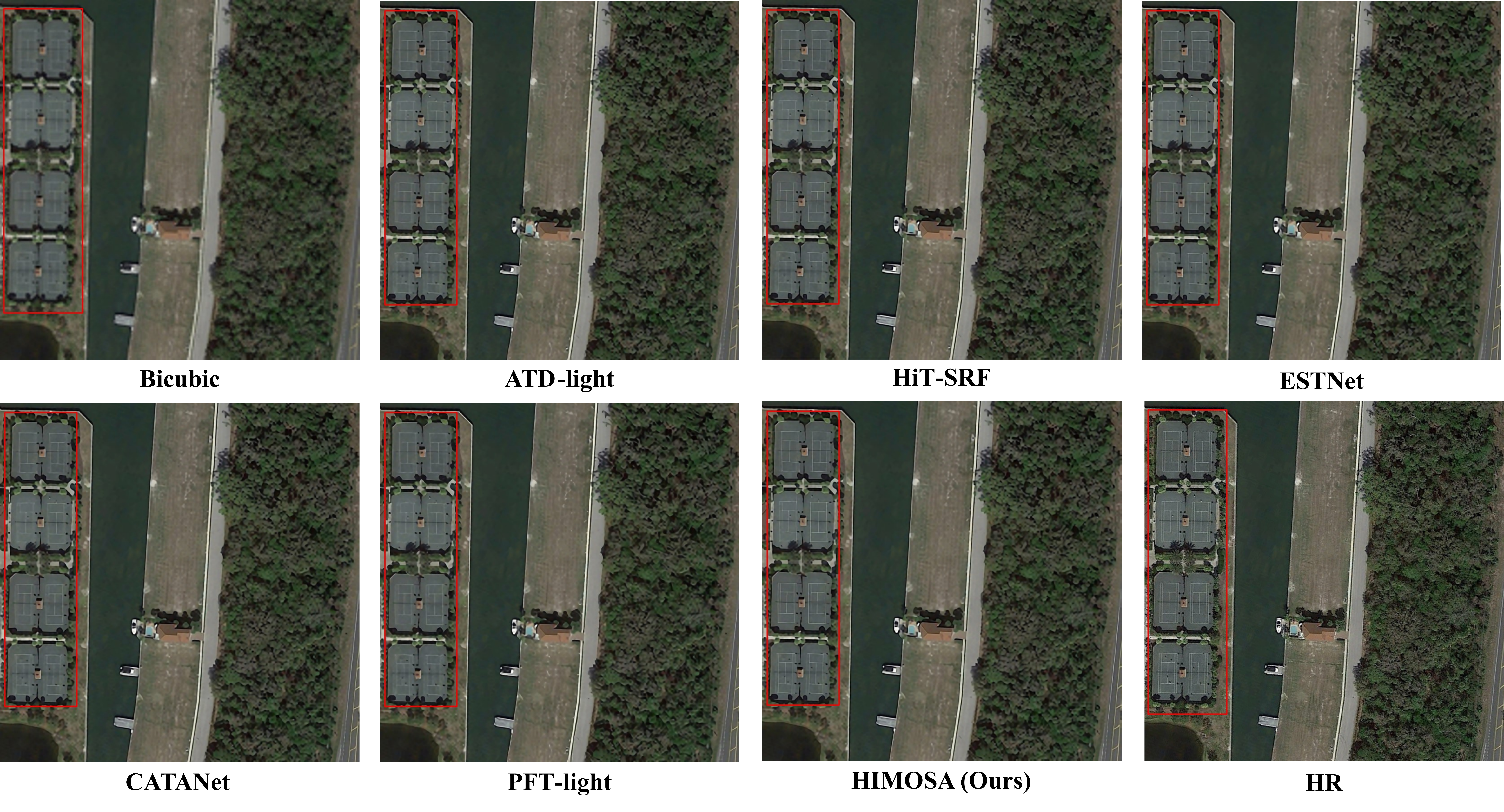}
\vspace{-4mm}
\caption{Visualization results ($\times 4$) of different methods in DIOR (zoom in for details).}
\vspace{-2mm}
\label{fig.comparison_dior}
\end{figure*}

\begin{figure*}[t]
\centering
\includegraphics[width=0.96  \textwidth]{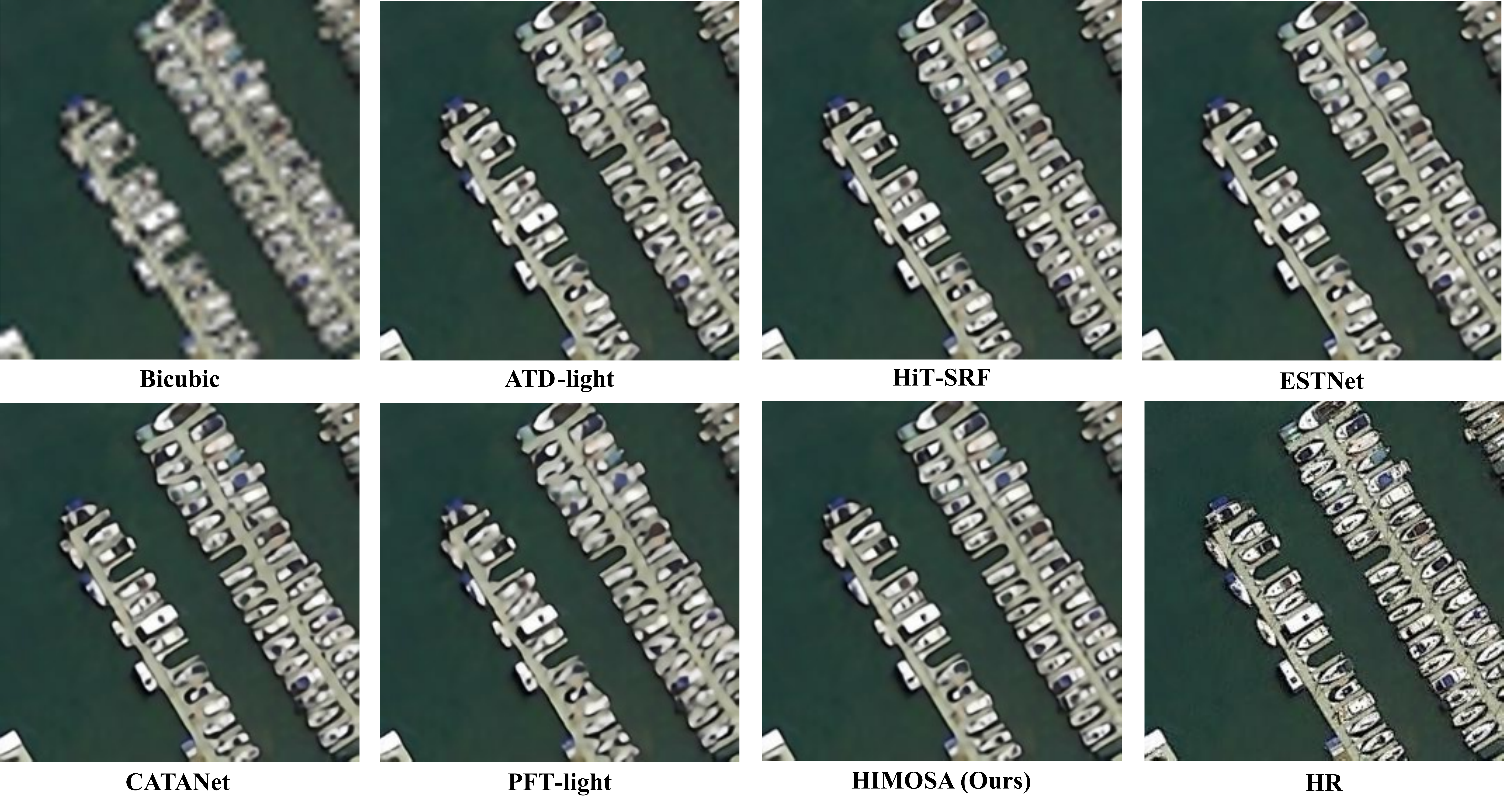}
\vspace{-4mm}
\caption{Visualization results ($\times 4$) of different methods in NWPU (zoom in for details).}
\vspace{-2mm}
\label{fig.comparison_nwpu}
\end{figure*}

To further demonstrate the qualitative advantages of our proposed HIMOSA framework, this section presents additional visual comparison results across various datasets and scenes. These visualizations showcase the ability of our method to reconstruct fine structural details, preserve edge sharpness, and restore realistic textures in complex remote sensing environments. Compared with existing approaches, HIMOSA generates visually clearer and more natural results, particularly in regions with dense textures or repetitive patterns. These results further confirm the strong reconstruction capability and generalization performance of our model in diverse real-world scenarios. The visual results are shown in \cref{fig.comparison_dior}, and \cref{fig.comparison_nwpu}

\begin{table*}
\begin{center}
\caption{Description of Different Datasets Composition.}
\vspace{-2mm}
\resizebox{\textwidth}{!}{
    \begin{tabular}{ccccccc}
       \toprule
       \textbf{Phase}&\textbf{Name} & \textbf{Used (Total) Samples} &  \textbf{Scene Classes} & \textbf{Spatial resolution($m$)}& \textbf{Image sizes} & \textbf{Usage} \\
       \midrule
       \midrule
       Train & AID & 3000 (10,000) & 30& $0.5\sim8$ & $600\times600$& Scene classification  \\
        \midrule
       \multirow{4}{*}{Test}  & AID & 900 (10,000) & 30& $0.5\sim8$ & $600\times600$& Scene classification  \\
       & NWPU-RESISC45 & 315 (31,500) & 45 & $0.2\sim30$ & $256\times256$ & Scene classification \\
        & DIOR & 1000 (23,463) & 20 & $0.5\sim30$ & $800\times800$ & Object Detection \\
        & DOTA & 900 (2,806) & 14 & $0.3\sim10$ & $800 \sim 4000$& Object Detection \\
       \bottomrule
    \end{tabular}
}
\label{table_datasets}
\vspace{-4mm}
\end{center}
\end{table*}

\begin{figure*}[t]
\centering
\includegraphics[width=0.96  \textwidth]{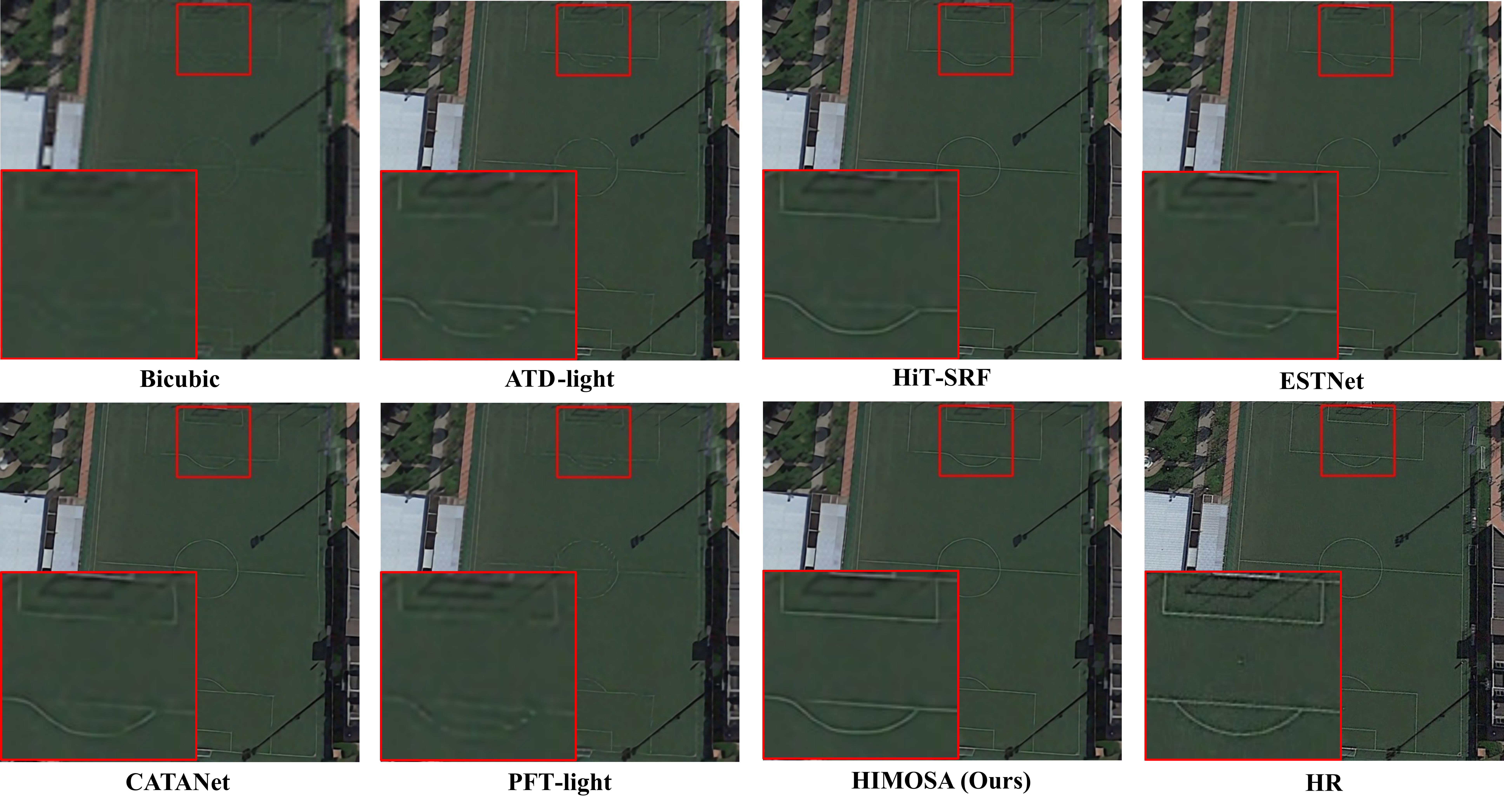}
\vspace{-4mm}
\caption{Visualization results ($\times 4$) achieved by different methods in AID datasets (zoom in for details).}
\vspace{-2mm}
\label{fig.comparison_aid}
\end{figure*}

\end{document}